\documentclass[11pt,a4paper]{article}
\usepackage{acl2022}
\usepackage{times}
\usepackage{latexsym}
\usepackage{csquotes}
\usepackage{url}
\usepackage{graphicx}
\usepackage{multicol,array,booktabs}
\usepackage{xcolor}
\usepackage{subcaption}
\usepackage{caption}
\usepackage{lipsum}
\usepackage{enumerate}
\usepackage{multirow}
\definecolor{markdap}{HTML}{730099}

\usepackage{microtype}

\title{Parsing linearizations appreciate PoS tags - but some are fussy about errors}

\author {\textbf{Alberto Muñoz-Ortiz$^1$}, \textbf{Mark Anderson$^{2}$}, \textbf{David Vilares$^1$}, \textbf{Carlos Gómez-Rodríguez$^1$} \\
$^1$Universidade da Coruña, CITIC, Spain\\
$^2$ PIN Caerdydd, Prifysgol Caerdydd, United Kingdom\\
\texttt{alberto.munoz.ortiz@udc.es, andersonm8@caerdydd.ac.uk,}\\ \texttt{david.vilares@udc.es, carlos.gomez@udc.es} \\
}
\date{}

\begin{document}
\maketitle
\begin{abstract}
PoS tags, once taken for granted as a useful resource for syntactic parsing, have become more situational with the popularization of deep learning. Recent work on the impact of PoS tags on graph- and transition-based parsers suggests that they are only useful when tagging accuracy is prohibitively high, or in low-resource scenarios. However, such an analysis is lacking for the emerging sequence labeling parsing paradigm, where it is especially relevant as some models explicitly use PoS tags for encoding and decoding. We undertake a study and uncover some trends. Among them, PoS tags are generally more useful for sequence labeling parsers than for other paradigms, but the impact of their accuracy is highly encoding-dependent, with the PoS-based head-selection encoding being best only when both tagging accuracy and resource availability are high.

\end{abstract}
\section{Introduction}
PoS tags have long been considered a useful feature for parsers, especially prior to the prevalence of neural networks \cite{voutilainen1998does,dalrymple2006much,alfared2012PoS}. For neural parsers, it is less clear if they are useful or not. Work has shown that when using word and character embeddings, PoS tags become much less useful \cite{ballesteros2015improved,de2017raw}. However, \citet{dozat2017stanford} found using universal PoS (UPoS) tags to be somewhat helpful, but improvements are typically quite small \cite{smith2018investigation}. Similarly, for multi-task systems, small improvements have been observed for both UPoS and finer-grained tags \cite{zhang2020PoS}.

A limiting factor when using predicted PoS tags is the apparent need for very high accuracy from taggers \cite{anderson-gomez-rodriguez-2020-frailty}. This is particularly problematic in a low-resource setting where using gold tags gives unreasonably high performance  \cite{tiedemann2015cross} and high accuracy taggers are difficult to obtain \cite{kann2020weakly}. However, some work has suggested that in a low-resource setting even low accuracy taggers can be beneficial for parsing performance, especially when there is more PoS tag annotations than dependency tree annotations \cite{anderson-etal-2021-falta}.

These findings relate to transition-based (TB) and graph-based (GB) parsers, but recently several encodings have been proposed to frame dependency parsing as a sequence labeling task~\citep{strzyz-etal-2019-viable,lacroix-2019-dependency,gomez-rodriguez-etal-2020-unifying}, providing an alternative to GB and TB models when efficiency is a priority~\citep{anderson-gomez-rodriguez-2021-modest}. \citet{munoz-ortiz-etal-2021-linearizations} found that the amount of data required for different encodings varied and that some were impacted by predicted PoS tag use more than others.

Here, we evaluate the impact of PoS tagging accuracy on different encodings and also the interplay of this potential relation and the amount of available data (using low-, mid-, high-, and very-high- resource treebanks). This is done by artificially controlling the accuracy of PoS taggers by using the nature of errors generated by robust taggers.\footnote{All source code available at \url{https://www.grupolys.org/software/aacl2022/}.}

\section{Sequence labeling parsing}
In dependency parsing as sequence labeling, the goal is to assign a single label of the form $(x_i,l_i)$ to every input token $w_i$ of a sequence, where $x_i$ encodes a subset of the arcs related to $w_i$ and $l_i$ is the dependency type. Below, we review the existing families of linearizations used in this work.\smallskip

\noindent\textbf{Head-selection} \cite{spoustova2010dependency}, where $x_i$ encodes the head of $w_i$ using an absolute index or a relative offset, that can be based on some word property (usually PoS tags, which is also the property we use in this work due to its strong performance in previous work). So for instance, if $x_i$ = (+\emph{n}, \textsc{x}), this would indicate that the head of $w_i$ is the $n$th word to the right of $w_i$ with the word property \textsc{x}. Some desirable properties of this encoding family are a direct correspondence between words and arcs and the capacity to encode any non-projective tree. However, a major weakness is its dependency on the chosen property (in our case, PoS tags) to decode trees.\smallskip

\noindent\textbf{Bracketing-based} $x_i$ represents the dependency arcs using a string of brackets, with each arc represented by a bracket pair. Its main advantage is that it is independent of external features, but regarding projectivity it cannot represent arcs that cross in the same direction. To alleviate this, we use the encoding proposed by \citet{strzyz-etal-2020-bracketing}, that adds a second independent plane of brackets (\texttt{2p\textsuperscript{b}}), inspired by multiplanarity \cite{yli2003multiplanarity}.\smallskip

\noindent \textbf{Transition-based} \cite{gomez-rodriguez-etal-2020-unifying}, wheregiven a sequence of transitions generated by a left-to-right transition-based parser, it splits it in labels based on read transitions (e.g. \textsc{shift}), such that each word receives a label $x_i$ with a subset of transition actions. For this work, we consider mappings from a projective algorithm, arc-hybrid \citep[\texttt{ah\textsuperscript{tb}};][]{kuhlmann-etal-2011-dynamic} and a non projective algorithm, Covington \citep[\texttt{c\textsuperscript{tb}};][]{covington2001fundamental}.

\subsection{Parser systems}
We use a 2-layer bidirectional long short-term memory (biLSTM) network with a feed-forward network to predict the labels using softmaxes. We use hard-sharing multi-task learning to predict $x_i$ and $l_i$.\footnote{We use a 2-task setup for all encodings, except  \texttt{2p\textsuperscript{b}} for which we use 3 tasks, as each plane is predicted independently.}
The inputs to the network are randomly initialized word embeddings and LSTM character embeddings and \emph{optionally} (see \S \ref{sec:experiments}), PoS tag embeddings. The appendix specifies the hyperparameters. For a homogeneous comparison against work on the usefulness of PoS tags for transition and graph-based models, and focused on efficiency, we do not use large language models.

\section{Controlling PoS tag accuracy}\label{sec:error_algorithm}
We purposefully change the accuracy of the PoS tags in a treebank, effectively treating this accuracy as the independent variable in a controlled experiment and LAS as the dependent variable, i.e. $LAS = f(Acc_{PoS})$ where $f$ is some function. Rather than randomly altering the gold label of PoS tags, we alter them based on the actual errors that PoS taggers make for a given treebank. This means PoS tags that are more likely to be incorrect for a given treebank will be more likely to be altered when changing the overall PoS accuracy of that treebank. We refer to this as the \textit{error rate} for PoS tags. The incorrect label is also based on the most likely incorrect label for the PoS tag error for that treebank based on the incorrect labeling from the tagger. We refer to this as the \textit{error type}, e.g. \texttt{NOUN}$\rightarrow$\texttt{VERB}.

We trained BiLSTM taggers for each of the treebanks to get the error rates for each PoS tag type and rate of each error type for each tag. Their generally high performances, even for the smaller treebanks, are shown in Table \ref{tab:tagger_acc} in the Appendix.

From the errors of these taggers, we first need the estimated probability that a given PoS tag $t$ is tagged erroneously:
\begin{equation}
    p(\mathit{error}|t) = \frac{E_{t}}{C_{t}}
\end{equation}
where $E_{t}$ is the error count for tag $t$ and $C_{t}$ is the total count for tag $t$. Then we need the probability of applying an erroneous tag $e$ to a ground-truth tag $t$:
\begin{equation}
    p(e|t,\mathit{error}) = \frac{E_{t\rightarrow e}}{E_{t}}
\end{equation}
where $E_{t\rightarrow e}$ is the error count when labeling $t$ as $e$. This estimated probability remains fixed, whereas $p(\mathit{error}|t)$ is adjusted to vary the overall accuracy.

We adjust these values by applying a weight, $\gamma$:
\begin{equation}
    \gamma = \frac{E_{A}}{E}
\end{equation}
where $E$ is the global error count and $E_{A}$ is the adjusted global error count such that the resulting tagging error is $A$. $p(\mathit{error}|t)$ is then adjusted:
\begin{equation}
    p(\mathit{error}|t) = \frac{\gamma E_{t}}{C_{t}}
\end{equation}
It is possible that $\gamma E_{t} > C_{t}$. When this occurs to tag $t$ we cap $\gamma E_{t}$ at $C_{t}$ and then recalculate $\gamma$, removing the counts associated with this tag:
\begin{equation}
    \gamma = \frac{E_{A}-C_{t}}{E-C_{t}}
\end{equation}
This is then done iteratively for each tag where $E_{t}\geq C_{t}$ until we obtain an error count for each tag such that the total error count reaches $E_A$.

These are all derived and applied as such to the test set of treebanks as this is where we evaluate the impact of PoS tag errors. To further echo the erroneous nature of these taggers, when $E_A\leq E$ only the subset of real errors are used when generating errors. When $E_A>E$ this subset of real errors is maintained and subtracted such that:
\begin{equation}
p(\mathit{error}|t) = \frac{(\gamma - 1)E_{t}}{C_{t}-E_{t}}
\end{equation}
and this is only applied on the tokens which were not erroneously tagged by the taggers.

\begin{table}[t!]
    \centering
    \scriptsize
    \begin{tabular}{cllrr}
    \toprule
    & Treebank & Family  & \# Trees & \# Tokens \\ \midrule
    \parbox[t]{2.5mm}{\multirow{4}{*}{\rotatebox[origin=c]{90}{\textsc{Low}}}} & Skolt Sami\textsubscript{Giellagas} & Uralic (Sami) & 200 & 2\,461\\
    &Guajajara\textsubscript{TuDeT} & Tupian (Tupi-Guarani) & 284 & 2\,052\\
    &Ligurian\textsubscript{GLT} & IE (Romance) & 316 & 6\,928 \\
    &Bhojpuri\textsubscript{BHTB} & IE (Indic) & 357 & 6\,665 \\
    \midrule
    \parbox[t]{2.5mm}{\multirow{4}{*}{\rotatebox[origin=c]{90}{\textsc{Mid}}}}&Kiche\textsubscript{IU} & Mayan & 1\,435 & 10\,013 \\
    &Welsh\textsubscript{CCG} & IE (Celtic) & 2\,111 &  41\,208 \\
    &Armenian\textsubscript{ArmTDP} & IE (Armenian) & 2\,502 & 52\,630 \\
    &Vietnamese\textsubscript{VTB} & Austro-Asiatic (Viet-Muong) & 3\,000 & 43\,754 \\
    \midrule
    \parbox[t]{2.5mm}{\multirow{4}{*}{\rotatebox[origin=c]{90}{\textsc{High}}}}&Basque\textsubscript{BDT} & Basque & 8\,993 & 121\,443 \\
    &Turkish\textsubscript{BOUN} & Turkic (Southwestern) & 9\,761 & 122\,383 \\
    &Bulgarian\textsubscript{BTB} & IE (Slavic) & 11\,138 & 146\,159 \\
    &Ancient Greek\textsubscript{Perseus} & IE (Greek) & 13\,919 & 202\,989 \\
    \midrule
    \parbox[t]{2.5mm}{\multirow{4}{*}{\rotatebox[origin=c]{90}{\textsc{V. High}}}} & Norwegian\textsubscript{Bokmål} & IE (Germanic) & 20\,044 & 310\,221 \\
    & Korean\textsubscript{Kaist} & Korean & 27\,363 & 350\,090 \\
    & Persian\textsubscript{PerDT }& IE (Iranian) & 29\,107 & 501\,776 \\
    &Estonian\textsubscript{EDT} & Uralic (Finnic) & 30\,972 & 437\,769 \\
    \bottomrule
    \end{tabular}
    \caption{Details of the treebanks used in this work.}
    \label{tab:tb_info}
\end{table}

\begin{table*}[thbp!]
    \centering
    \scriptsize
    \begin{tabular}{l|llll|llll|llll|llll||llll}
    \toprule
    \multirow{2}{*}{Setup}& \multicolumn{4}{c|}{Low-resource} & \multicolumn{4}{c|}{Mid-resource} & \multicolumn{4}{c|}{High-resource} & \multicolumn{4}{c||}{V. high-resource} & \multicolumn{4}{c}{All} \\
     & \texttt{2p\textsuperscript{b}} & \texttt{ah\textsuperscript{tb}} & \texttt{c\textsuperscript{tb}} & \texttt{rp\textsuperscript{h}}
     & \texttt{2p\textsuperscript{b}} & \texttt{ah\textsuperscript{tb}} & \texttt{c\textsuperscript{tb}} & \texttt{rp\textsuperscript{h}}
     & \texttt{2p\textsuperscript{b}} & \texttt{ah\textsuperscript{tb}} & \texttt{c\textsuperscript{tb}} & \texttt{rp\textsuperscript{h}}
     & \texttt{2p\textsuperscript{b}} & \texttt{ah\textsuperscript{tb}} & \texttt{c\textsuperscript{tb}} & \texttt{rp\textsuperscript{h}}
     & \texttt{2p\textsuperscript{b}} & \texttt{ah\textsuperscript{tb}} & \texttt{c\textsuperscript{tb}} & \texttt{rp\textsuperscript{h}}\\
    \midrule
     75 & \textbf{50.65} & 49.33 & 48.43 & 47.72 & \textbf{63.26} & 60.18 & 60.23 & 58.64 & \textbf{66.34} & 64.18 & 63.87 & 64.09 & \textbf{79.63} & 77.44 & 75.26 & 73.32 & \textbf{64.97} & 62.78 & 61.98 & 60.94 \\
     80 & \textbf{53.84} & 50.58 & 48.78 & 50.94 & \textbf{64.00} & 61.52 & 61.34 & 60.87 & \textbf{67.53} & 64.88 & 64.88 & 64.70 & \textbf{80.06} & 77.93 & 75.74 & 77.09 & \textbf{66.36} & 63.73 & 62.69 & 63.40 \\ 
     85 & \textbf{54.17} & 52.48 & 51.27 & 52.62 & \textbf{65.25} & 62.34 & 62.06 & 63.36 & \textbf{68.11} & 65.38 & 65.33 & 66.56 & \textbf{81.18} & 79.02 & 77.34 & 78.76 & \textbf{67.18} & 64.81 & 64.00 & 65.32\\
     90 & \textbf{56.03} & 53.55 & 52.78 & 55.34 & \textbf{67.30} & 64.05 & 63.35 & 66.18 & 69.31 & 66.86 & 66.61 & \textbf{69.47} & \textbf{81.33} & 79.39 & 77.05 & 79.80 & \textbf{68.49} & 65.96 & 65.01 & 67.70\\
     95 & \textbf{59.30} & 56.88 & 55.75 & 58.90 & 69.84 & 67.34 & 66.20 & \textbf{70.30} & 70.28 & 67.66 & 67.32 & \textbf{71.18} & \textbf{82.61} & 80.62 & 78.83 & 82.52 & 70.51 & 68.12 & 67.02 & \textbf{70.72}\\
     97.5 & 60.00 & 58.70 & 57.59 & \textbf{61.86} & 72.63 & 69.47 & 68.99 & \textbf{72.84} & 71.59 & 69.27 & 68.39 & \textbf{72.83} & 83.91 & 82.00 & 80.27 & \textbf{84.31} & 71.96 & 69.86 & 68.81 & \textbf{72.96}\\
     100 & 62.16 & 60.97 & 58.64 & \textbf{64.23} & 74.28 & 71.19 & 70.02 & \textbf{75.20} & 73.40 & 70.60 & 70.05 & \textbf{74.50} & 86.52 & 84.77 & 82.65 & \textbf{87.20} & 74.09 & 71.88 & 70.34 &\textbf{75.24}\\
     MTL & 47.78 & 46.83 & 45.60 & \textbf{48.08} & \textbf{64.15} & 62.15 & 60.68 & 63.17 & \textbf{67.97} & 64.94 & 65.26 & 67.47 & \textbf{81.52} & 79.46 & 76.85 & 80.95 & \textbf{65.35} & 63.34 & 62.10 & 64.92 \\
     No PoS tags & 47.36 & 46.18 & 45.79 & \textbf{49.26} & \textbf{63.94} & 61.58 & 60.73 & 57.52 & \textbf{67.67} & 64.76 & 64.75 & 66.58 & \textbf{81.15} & 79.22 & 76.98 & 80.06 & \textbf{65.03} & 62.94 & 62.06 & 63.35\\
    \bottomrule
    \end{tabular}
    \caption{Average LAS for different setups and PoS tag accuracies for the groups of treebanks studied.}
    \label{tab:scores}
\end{table*}

For every eligible token, based on its tag $t$ an error is generated based on $p(\mathit{error}|t)$ and if an error is to be generated, the erroneous tag is selected based on the distribution over $p(e|t,\mathit{error})$.

This is also applied to the training and dev set as it seems better to use predicted tags when training \cite{anderson-gomez-rodriguez-2020-frailty}. There are differences in the distribution of PoS tags and as the algorithm is based on the test data, at times it isn't possible to get exactly $E_A$. We therefore allow a small variation of $\pm$0.05 on $E_A$.

We then selected a set of PoS tag accuracies to test a range of values (75, 80, 85, 95, 97.5, 100). We included the 97.5\% accuracy to evaluate the findings of \citet{anderson-gomez-rodriguez-2020-frailty}, where they observed a severe increase in performance between high scoring taggers and gold tags, otherwise we use increments of 5\%.

\section{Experiments}\label{sec:experiments}
We now present the experimental setup to determine how parsing scores evolve for the chosen linearizations when the tagging accuracy degrades. As evaluation metrics, we use Labeled (LAS) and Unlabeled Attachment Scores (UAS).

\paragraph{Data}
Treebanks from Table \ref{tab:tb_info} were selected using a number of criteria. We chose treebanks that were all from different language families and therefore exhibit a range of linguistic behaviors. We also selected treebanks such that we used 4 low-resource, 4 mid-resource, 4 high-resource and 4 very high-resource treebanks. Within each of those categories, we also selected treebanks with slightly different amounts of data, so as to obtain an incremental range of treebank sizes across low, mid, high and very high boundaries. Moreover, we ensured the quality of the treebanks by selecting treebanks that were either manually annotated in the UD framework or manually checked after automatic conversions. When a treebank did not contain a development set, we re-split the data by collecting the data across the training and test data and split the full data such that 60\% was allocated to the training set, 10\% to the development, and 30\% to the test.

\paragraph{Setup} We train and test parsers on sets of predicted tags, as explained in \S \ref{sec:error_algorithm}.We consider two baselines: (i) parsers trained without PoS tags\footnote{Forced setup for \texttt{rp\textsuperscript{h}}, as PoS tags are needed to decode.} (\texttt{base-no-tags}), (ii) parsers trained with gold tags on a multi-task setup (\texttt{base-mtl}).

\begin{figure}[t!]
    \centering
    \includegraphics[width=0.40\textwidth]{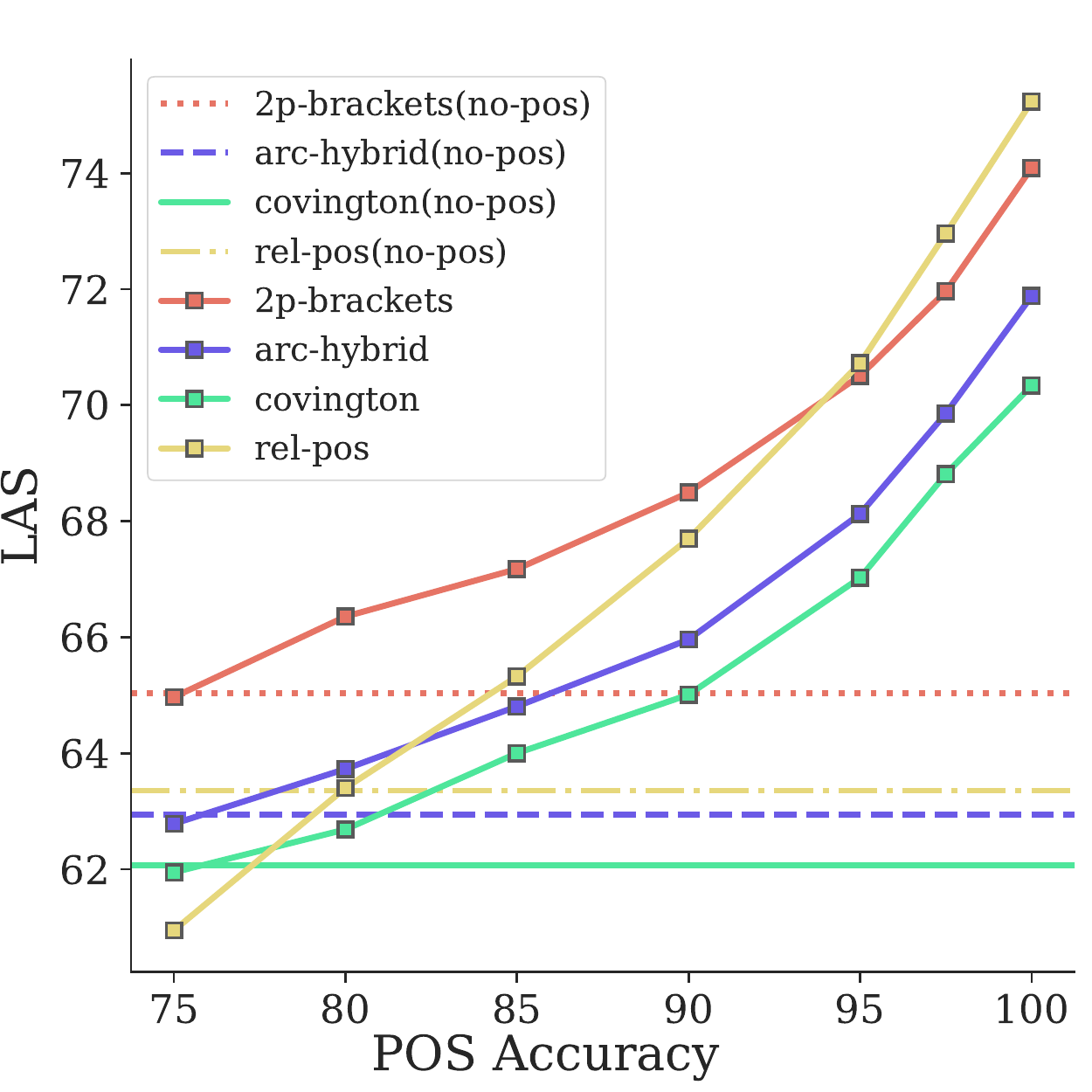}
    \caption{Average LAS across all treebanks against PoS tagging accuracies for different linearizations, compared to the no-tags baselines.}
    \label{fig:all_scores}
\end{figure}

\begin{figure*}
    \centering
    \begin{subfigure}{0.4\textwidth}
    \includegraphics[width=\textwidth]{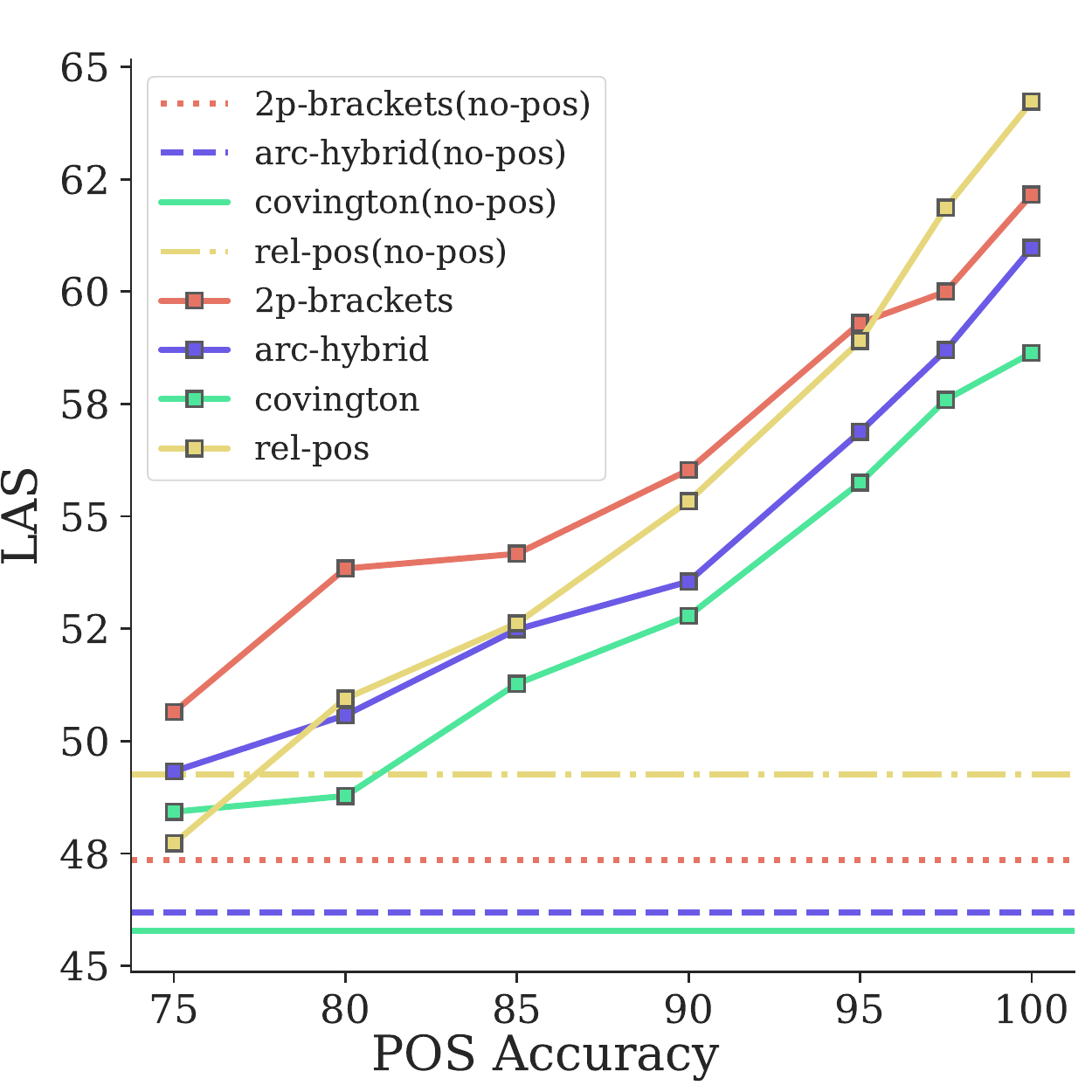}
    \caption{Low}
    \label{fig:low_split}
    \end{subfigure}
    ~
    \begin{subfigure}{0.4\textwidth}
    \includegraphics[width=\textwidth]{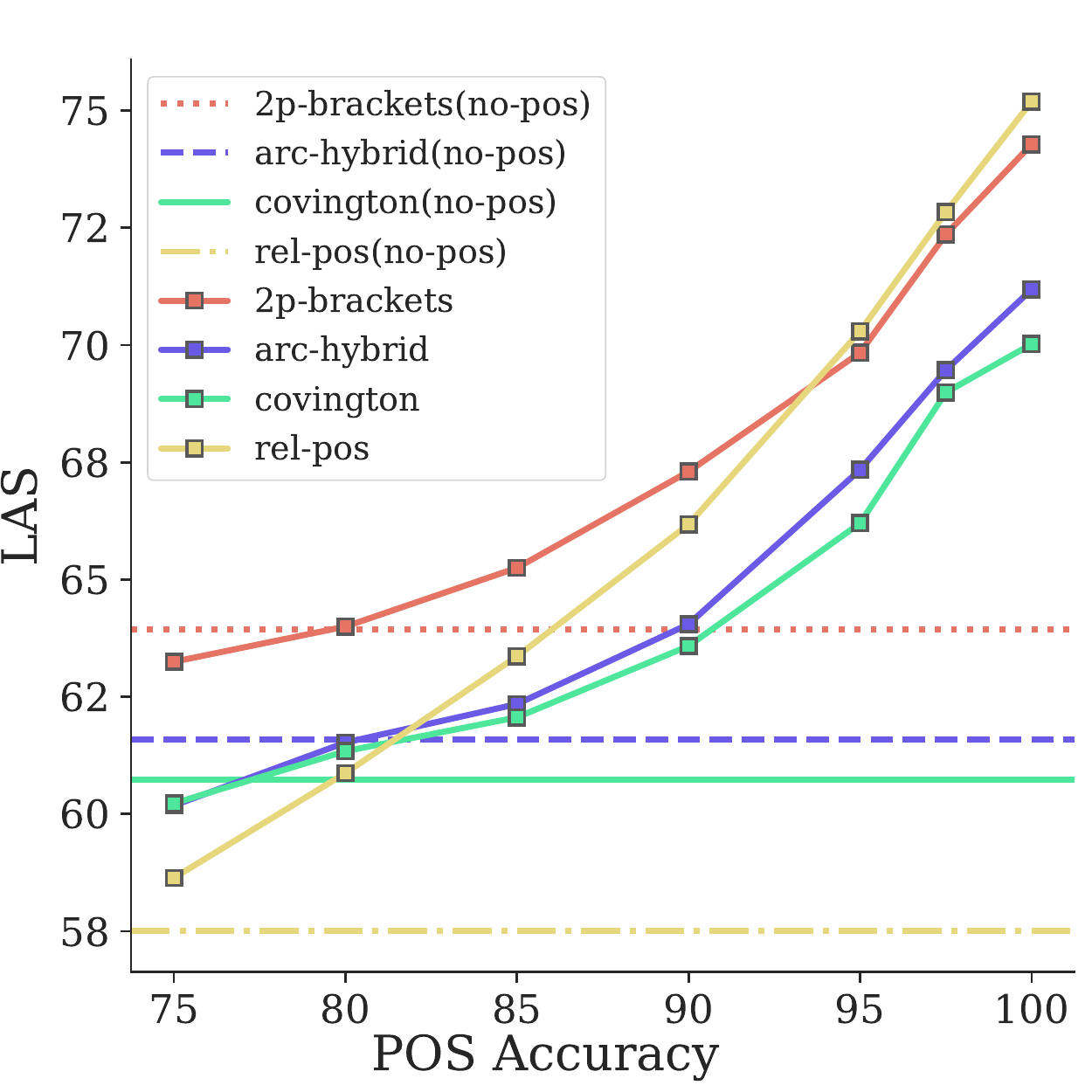}
    \caption{Mid}
    \label{fig:mid_split}
    \end{subfigure}
    
    \begin{subfigure}{0.4\textwidth}
    \includegraphics[width=\textwidth]{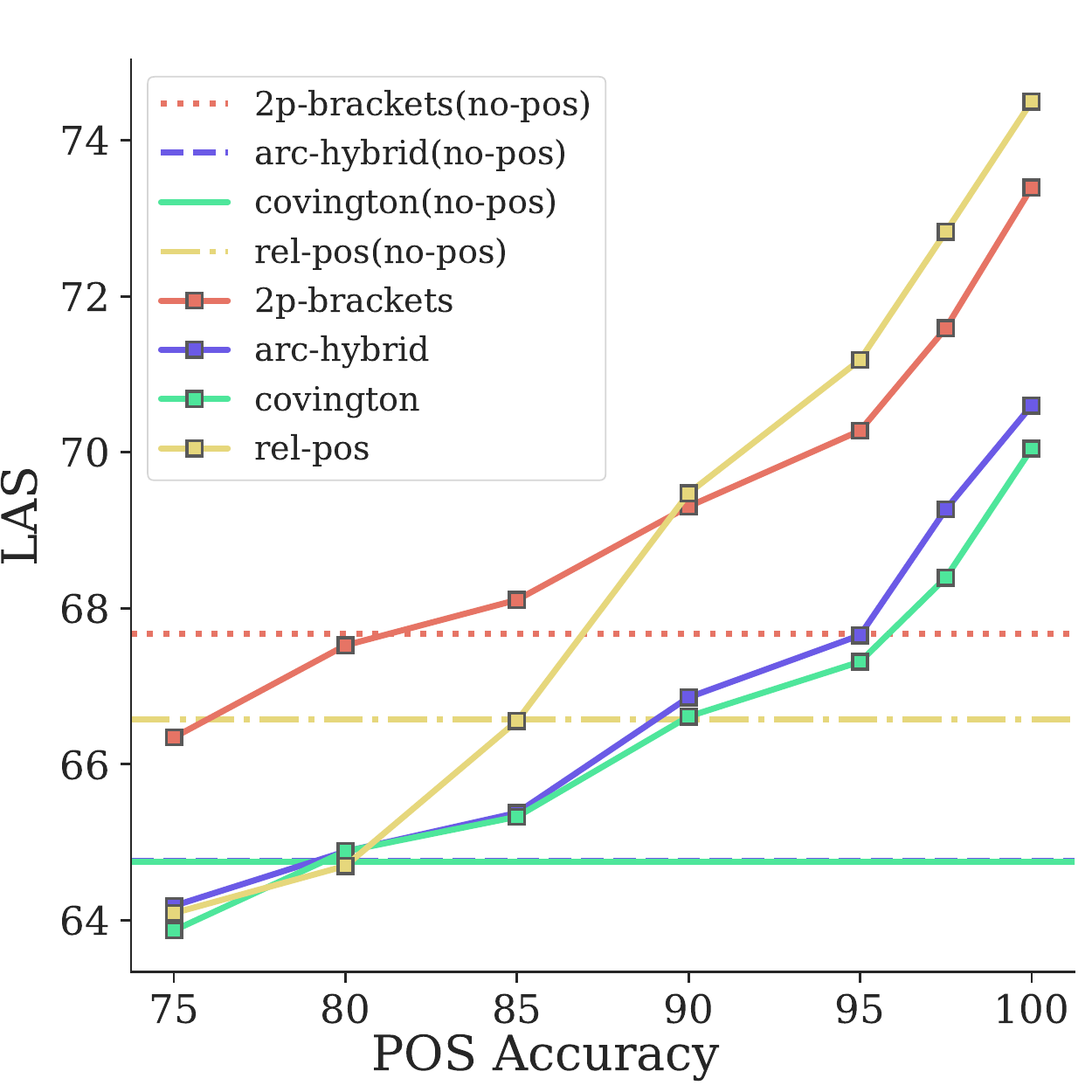}
    \caption{High}
    \label{fig:rich_split}
    \end{subfigure}
    ~
    \begin{subfigure}{0.4\textwidth}
    \includegraphics[width=\textwidth]{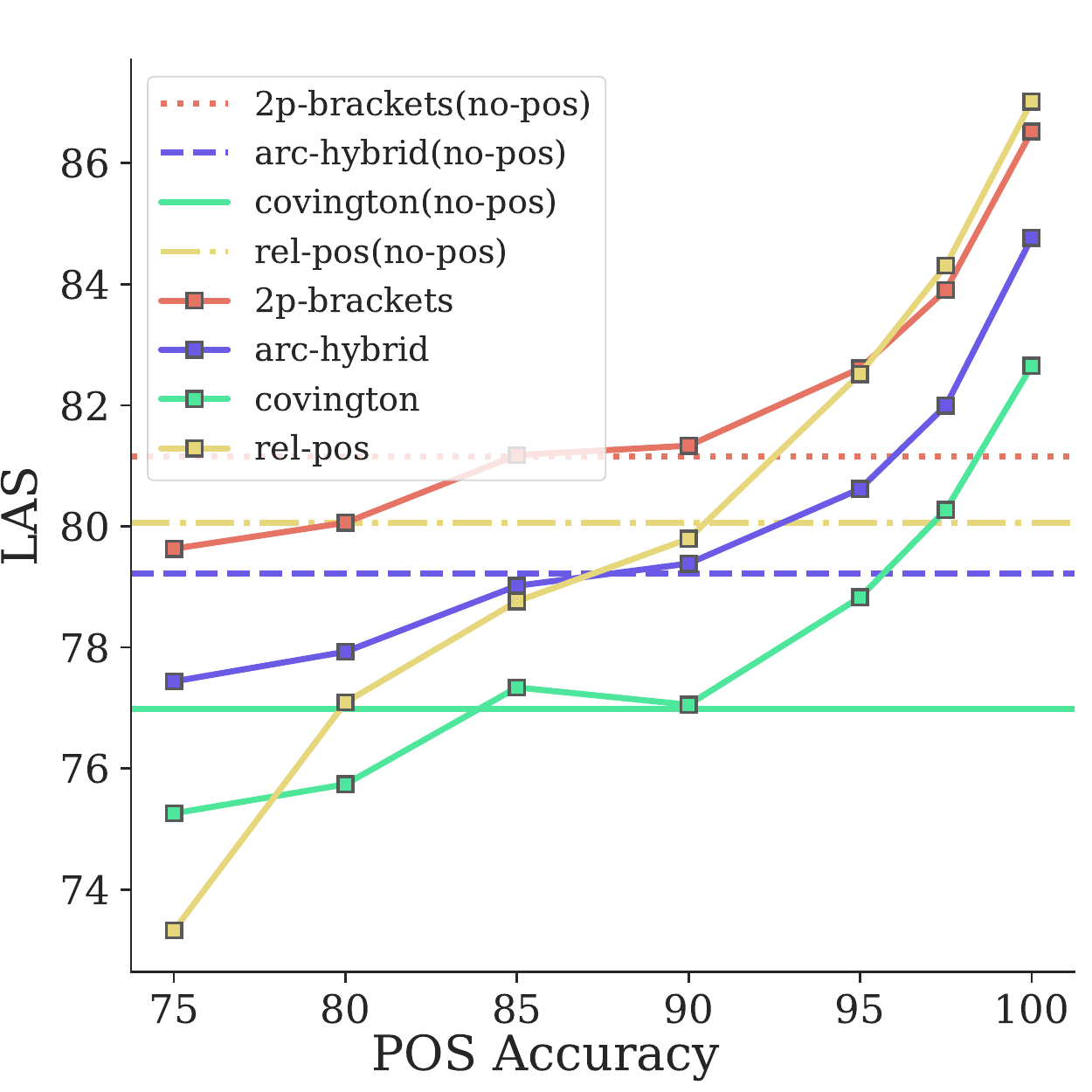}
    \caption{Very high}
    \label{fig:vhigh_split}
    \end{subfigure}
    \caption{Average LAS for the (a) low-, (b) mid-, (c) high- and (d) very high-resource subsets of treebanks for different PoS tagging accuracies and linearizations, compared to the no-tags baselines.}
    \label{fig:split_scores}
\end{figure*}

\subsection{Results}
Table \ref{tab:scores} shows the average LAS scores across all treebank setups for all encodings and tagging accuracies, together with both baselines. To better interpret the results and tendencies, we will also visualize the results in different figures. \footnote{UAS results are shown in Figures \ref{fig:all_scores_uas} and \ref{fig:split_scores_uas} in the Appendix.} Note that we don't include \texttt{base-mtl} as they performed very similar to \texttt{base-no-tags}. We include the results with a state-of-the-art graph based parser \cite{dozat2017stanford} in Table \ref{tab:supar_scores} for comparison.

\begin{table}[thbp!]
    \centering
    \scriptsize
    \begin{tabular}{l|llll||l}
    \toprule
    Setup & Low & Mid & High & V. High & All\\
    \midrule
     75 & 55.61 & 69.79 & 76.66 & 86.00 & 72.01\\
     80 & 56.60 & 70.17 & 76.49 & 85.95 & 72.30\\
     85 & 59.12 & 70.76 & 76.90 & 86.33 & 73.28\\
     90 & 60.40 & 71.61 & 77.69 & 86.62 & 74.08\\
     95 & 62.12 & 74.63 & 78.22 & 87.13 & 75.52\\
     97.5 & 65.05 & 76.42 & 79.44 & 88.16 & 77.27\\
     100 & 66.65 & 78.52 & 80.96 & 90.74 & 79.22\\
     No PoS tags & 58.40 & 71.71 & 77.66 & 87.72 & 73.74\\
    \bottomrule
    \end{tabular}
    \caption{Average LAS for different setups and PoS tag accuracies for the groups of treebanks studied using the graph-based parser.}
    \label{tab:supar_scores}
\end{table}

\paragraph{All treebanks}
Figure \ref{fig:all_scores} shows the average LAS across all treebanks for the different linearizations, using PoS tags or not. The results suggest that even using low accuracy tags  is better than not using them. In detail, \texttt{rp\textsuperscript{h}} is the linearization that is affected the most by the quality of the PoS tags, as it relies directly on them in order to decode the tree,  degrading from the 1st position when using gold tags to the last one when tags have an accuracy of 75\%. On the other hand, \texttt{2p\textsuperscript{b}} seems to be the most useful encoding for real-world situations, outperforming the other linearizations when no tags or tags with an accuracy under 95\% are used, and performing on par with \texttt{rp\textsuperscript{h}} over that mark. Note that while \citet{strzyz-etal-2019-viable} chose \texttt{rp\textsuperscript{h}} as their best model for evaluation, the choice was biased by using English, a language with atypically high tagging accuracy.

\paragraph{Results for different resourced sets of treebanks}
Figure \ref{fig:split_scores} shows the results for the low-resource, mid-resource, high-resource and very high-resource treebanks, respectively. Interestingly, we observe trends regarding the \emph{cutoff points} (the points where a model surpasses another), depending on the quality of PoS tags and quantity of available data. In particular, the cutoff points between the parsers that use PoS tags and the \texttt{base-no-tags} models are found at higher tagging accuracies when the data resources are larger too. Also, the cutoff point between \texttt{rp\textsuperscript{h}} and \texttt{2p\textsuperscript{b}} is at a lower PoS tagging accuracy when we have more data, although the results for the very high-resource treebanks break this trend. Finally, the low performance of the transition-based encodings is more pronounced for high-resource treebanks, with the exception the \texttt{ah\textsuperscript{tb}} for the very high-resource treebanks.

\section{Discussion}
The obtained results offer some valuable information about how PoS tag quality affects performance for different encodings and quantities of data. In most situations using PoS tags as features is better than not using them, in contrast with results for other parser architectures as described above.

In addition, the less resources, the harder it is for \texttt{rp\textsuperscript{h}} to beat brackets: cutoffs are at 97.5\%, 95\%, 90\% for low-, mid-, and high-resource treebanks, respectively. However, for very high-resource treebanks, the cutoff is back at 95\%. Compounded with the low tagging accuracy expected in low-resource setups, this highlights that \texttt{rp\textsuperscript{h}} is less suited for them. \texttt{2p\textsuperscript{b}}, which generally outperforms the other encodings below 90\% tagging accuracy, is the best low-resource option.

The more resources available, the harder it is for the models using PoS tags to outperform  \texttt{base-no-tags}, both for bracketing- and transition-based linearizations; i.e. experiments suggest that the benefits provided by the PoS tags decline when more training data is available. For brackets, the cutoffs occur at $<$75\%, 80\%, 85\% and 90\% for the low-, mid-, high- and very high-resource set, and for transition encodings,they are at $<$75\% for the low-resource set and at $\sim$80\% for mid- and high-resource sets. For the very-high resource set, cutoff points are at 85\% for \texttt{c\textsuperscript{tb}} and 90\% for \texttt{ah\textsuperscript{tb}}. 

\section{Conclusion}
We connected the impact that the quality of PoS tags and quantity of available data has
on several dependency parsing linearizations. We tested this by controlling PoS tagging performance on a range of UD treebanks, diverse in terms of both amount of resources and typology. The results showed that for sequence labeling parsing, which prioritizes efficiency, PoS tags are still welcome, contrary to more mature parsing paradigms such as transition-based and graph-based ones. The experiments also showed that parsing linearizations benefit from PoS tagging accuracy differently, and in particular linearizations that represent arcs as bracket strings are a better choice for most realistic scenarios.

\section*{Aknowledgements}
Mark was supported by a UKRI Future Leaders Fellowship (MR/T042001/1). This paper has received funding from ERDF/MICINN-AEI (SCANNER-UDC, PID2020-113230RB-C21), Xunta de Galicia (ED431C 2020/11), and Centro de Investigaci\'on de Galicia ``CITIC'', funded by Xunta de Galicia and the European Union (ERDF - Galicia 2014-2020 Program), by grant ED431G 2019/01

\bibliography{acl2021, anthology}

\begin{thebibliography}{22}
\expandafter\ifx\csname natexlab\endcsname\relax\def\natexlab#1{#1}\fi

\bibitem[{Alfared and B{\'e}chet(2012)}]{alfared2012PoS}
Ramadan Alfared and Denis B{\'e}chet. 2012.
\newblock {POS} taggers and dependency parsing.
\newblock \emph{International Journal of Computational Linguistics and
  Applications}, 3(2):107--122.

\bibitem[{Anderson et~al.(2021)Anderson, Dehouck, and
  G{\'o}mez-Rodr{\'\i}guez}]{anderson-etal-2021-falta}
Mark Anderson, Mathieu Dehouck, and Carlos G{\'o}mez-Rodr{\'\i}guez. 2021.
\newblock \href {https://doi.org/10.18653/v1/2021.iwpt-1.8} {A falta de pan,
  buenas son tortas: The efficacy of predicted {UPOS} tags for low resource
  {UD} parsing}.
\newblock In \emph{Proceedings of the 17th International Conference on Parsing
  Technologies and the IWPT 2021 Shared Task on Parsing into Enhanced Universal
  Dependencies (IWPT 2021)}, pages 78--83, Online. Association for
  Computational Linguistics.

\bibitem[{Anderson and
  G{\'o}mez-Rodr{\'\i}guez(2020)}]{anderson-gomez-rodriguez-2020-frailty}
Mark Anderson and Carlos G{\'o}mez-Rodr{\'\i}guez. 2020.
\newblock \href {https://doi.org/10.18653/v1/2020.conll-1.6} {On the frailty of
  universal {POS} tags for neural {UD} parsers}.
\newblock In \emph{Proceedings of the 24th Conference on Computational Natural
  Language Learning}, pages 69--96, Online. Association for Computational
  Linguistics.

\bibitem[{Anderson and
  G{\'o}mez-Rodr{\'\i}guez(2021)}]{anderson-gomez-rodriguez-2021-modest}
Mark Anderson and Carlos G{\'o}mez-Rodr{\'\i}guez. 2021.
\newblock \href {https://doi.org/10.18653/v1/2021.iwpt-1.12} {A modest {P}areto
  optimisation analysis of dependency parsers in 2021}.
\newblock In \emph{Proceedings of the 17th International Conference on Parsing
  Technologies and the IWPT 2021 Shared Task on Parsing into Enhanced Universal
  Dependencies (IWPT 2021)}, pages 119--130, Online. Association for
  Computational Linguistics.

\bibitem[{Ballesteros et~al.(2015)Ballesteros, Dyer, and
  Smith}]{ballesteros2015improved}
Miguel Ballesteros, Chris Dyer, and Noah~A Smith. 2015.
\newblock Improved transition-based parsing by modeling characters instead of
  words with {LSTM}s.
\newblock \emph{arXiv preprint arXiv:1508.00657}.

\bibitem[{Covington(2001)}]{covington2001fundamental}
Michael~A. Covington. 2001.
\newblock A fundamental algorithm for dependency parsing.
\newblock In \emph{Proceedings of the 39th annual ACM southeast conference},
  volume~1. Citeseer.

\bibitem[{Dalrymple(2006)}]{dalrymple2006much}
Mary Dalrymple. 2006.
\newblock How much can part-of-speech tagging help parsing?
\newblock \emph{Natural Language Engineering}, 12(4):373--389.

\bibitem[{de~Lhoneux et~al.(2017)de~Lhoneux, Shao, Basirat, Kiperwasser,
  Stymne, Goldberg, and Nivre}]{de2017raw}
Miryam de~Lhoneux, Yan Shao, Ali Basirat, Eliyahu Kiperwasser, Sara Stymne,
  Yoav Goldberg, and Joakim Nivre. 2017.
\newblock From raw text to universal dependencies-look, no tags!
\newblock In \emph{Proceedings of the CoNLL 2017 Shared Task: Multilingual
  Parsing from Raw Text to Universal Dependencies}, pages 207--217.

\bibitem[{Dozat et~al.(2017)Dozat, Qi, and Manning}]{dozat2017stanford}
Timothy Dozat, Peng Qi, and Christopher~D. Manning. 2017.
\newblock Stanford’s graph-based neural dependency parser at the {CoNLL} 2017
  shared task.
\newblock In \emph{Proceedings of the CoNLL 2017 Shared Task: Multilingual
  Parsing from Raw Text to Universal Dependencies}, pages 20--30.

\bibitem[{G{\'o}mez-Rodr{\'\i}guez et~al.(2020)G{\'o}mez-Rodr{\'\i}guez,
  Strzyz, and Vilares}]{gomez-rodriguez-etal-2020-unifying}
Carlos G{\'o}mez-Rodr{\'\i}guez, Michalina Strzyz, and David Vilares. 2020.
\newblock \href {https://doi.org/10.18653/v1/2020.coling-main.336} {A unifying
  theory of transition-based and sequence labeling parsing}.
\newblock In \emph{Proceedings of the 28th International Conference on
  Computational Linguistics}, pages 3776--3793, Barcelona, Spain (Online).
  International Committee on Computational Linguistics.

\bibitem[{{Kann} et~al.(2020){Kann}, {Lacroix}, and
  {S{\o}gaard}}]{kann2020weakly}
Katharina {Kann}, Oph\'{e}lie {Lacroix}, and Anders {S{\o}gaard}. 2020.
\newblock Weakly supervised {POS} taggers perform poorly on truly low-resource
  languages.
\newblock \emph{Proceedings of the AAAI Conference on Artificial Intelligence},
  34(5):8066--8073.

\bibitem[{Kuhlmann et~al.(2011)Kuhlmann, G{\'o}mez-Rodr{\'\i}guez, and
  Satta}]{kuhlmann-etal-2011-dynamic}
Marco Kuhlmann, Carlos G{\'o}mez-Rodr{\'\i}guez, and Giorgio Satta. 2011.
\newblock \href {https://aclanthology.org/P11-1068} {Dynamic programming
  algorithms for transition-based dependency parsers}.
\newblock In \emph{Proceedings of the 49th Annual Meeting of the Association
  for Computational Linguistics: Human Language Technologies}, pages 673--682,
  Portland, Oregon, USA. Association for Computational Linguistics.

\bibitem[{Lacroix(2019)}]{lacroix-2019-dependency}
Oph{\'e}lie Lacroix. 2019.
\newblock \href {https://doi.org/10.18653/v1/W19-7716} {Dependency parsing as
  sequence labeling with head-based encoding and multi-task learning}.
\newblock In \emph{Proceedings of the Fifth International Conference on
  Dependency Linguistics (Depling, SyntaxFest 2019)}, pages 136--143, Paris,
  France. Association for Computational Linguistics.

\bibitem[{Mu{\~n}oz-Ortiz et~al.(2021)Mu{\~n}oz-Ortiz, Strzyz, and
  Vilares}]{munoz-ortiz-etal-2021-linearizations}
Alberto Mu{\~n}oz-Ortiz, Michalina Strzyz, and David Vilares. 2021.
\newblock \href {https://aclanthology.org/2021.ranlp-1.111} {Not all
  linearizations are equally data-hungry in sequence labeling parsing}.
\newblock In \emph{Proceedings of the International Conference on Recent
  Advances in Natural Language Processing (RANLP 2021)}, pages 978--988, Held
  Online. INCOMA Ltd.

\bibitem[{Smith et~al.(2018)Smith, de~Lhoneux, Stymne, and
  Nivre}]{smith2018investigation}
Aaron Smith, Miryam de~Lhoneux, Sara Stymne, and Joakim Nivre. 2018.
\newblock An investigation of the interactions between pre-trained word
  embeddings, character models and {POS} tags in dependency parsing.
\newblock \emph{arXiv preprint arXiv:1808.09060}.

\bibitem[{Spoustov{\'a} and Spousta(2010)}]{spoustova2010dependency}
Drahom{\'i}ra~Johanka Spoustov{\'a} and Miroslav Spousta. 2010.
\newblock Dependency parsing as a sequence labeling task.
\newblock \emph{The Prague Bulletin of Mathematical Linguistics},
  94(2010):7--14.

\bibitem[{Strzyz et~al.(2019)Strzyz, Vilares, and
  G{\'o}mez-Rodr{\'\i}guez}]{strzyz-etal-2019-viable}
Michalina Strzyz, David Vilares, and Carlos G{\'o}mez-Rodr{\'\i}guez. 2019.
\newblock \href {https://doi.org/10.18653/v1/N19-1077} {Viable dependency
  parsing as sequence labeling}.
\newblock In \emph{Proceedings of the 2019 Conference of the North {A}merican
  Chapter of the Association for Computational Linguistics: Human Language
  Technologies, Volume 1 (Long and Short Papers)}, pages 717--723, Minneapolis,
  Minnesota. Association for Computational Linguistics.

\bibitem[{Strzyz et~al.(2020)Strzyz, Vilares, and
  G{\'o}mez-Rodr{\'\i}guez}]{strzyz-etal-2020-bracketing}
Michalina Strzyz, David Vilares, and Carlos G{\'o}mez-Rodr{\'\i}guez. 2020.
\newblock \href {https://doi.org/10.18653/v1/2020.coling-main.223} {Bracketing
  encodings for 2-planar dependency parsing}.
\newblock In \emph{Proceedings of the 28th International Conference on
  Computational Linguistics}, pages 2472--2484, Barcelona, Spain (Online).
  International Committee on Computational Linguistics.

\bibitem[{{Tiedemann}(2015)}]{tiedemann2015cross}
J{\"o}rg {Tiedemann}. 2015.
\newblock Cross-lingual dependency parsing with universal dependencies and
  predicted pos labels.
\newblock \emph{Proceedings of the Third International Conference on Dependency
  Linguistics (Depling 2015)}, pages 340--349.

\bibitem[{Voutilainen(1998)}]{voutilainen1998does}
Atro Voutilainen. 1998.
\newblock Does tagging help parsing?: a case study on finite state parsing.
\newblock In \emph{Proceedings of the International Workshop on Finite State
  Methods in Natural Language Processing}, pages 25--36. Association for
  Computational Linguistics.

\bibitem[{Yli-Jyr{\"a}(2003)}]{yli2003multiplanarity}
Anssi~Mikael Yli-Jyr{\"a}. 2003.
\newblock Multiplanarity-a model for dependency structures in treebanks.
\newblock In \emph{TLT 2003, Proceedings of the Second Workshop on Treebanks
  and Linguistic Theories}. V{\"a}xj{\"o} University Press.

\bibitem[{Zhang et~al.(2020)Zhang, Li, Zhou, and Zhang}]{zhang2020PoS}
Yu~Zhang, Zhenghua Li, Houquan Zhou, and Min Zhang. 2020.
\newblock Is {POS} tagging necessary or even helpful for neural dependency
  parsing?
\newblock \emph{arXiv preprint arXiv:2003.03204}.

\end{thebibliography}
\clearpage

\appendix
\section{PoS tagging details}

Table \ref {tab:tagger_hyperparameters} details the hyperparameters used to train the taggers in this work.

\begin{table}[htbp!]
\footnotesize
    \centering
        \tabcolsep=.25cm  
    \begin{tabular}{l p{1.5em} r}
    \toprule
    \textbf{Hyperparameter} & & \textbf{Value}\\
    \midrule
         Word embedding dimensions& & 100\\
         Character embedding in && 32 \\
         Character embedding out && 100 \\
         Embedding dropout &  & 0.33 \\
         biLSTM layers && 3 \\
         biLSTM nodes && 400 \\
         biLSTM dropout && 0.33 \\
         MLP dimensions  &  & 512\\
         MLP layers && 1 \\
         Epochs&  & 200 \\
         Patience&& 10 \\
         training batch size && 32 \\
          learning rate&  & 0.002 \\
          $\beta_{1}$, $\beta_{2}$ && 0.9, 0.9\\
          $\epsilon$ && $1\times10^{-12}$ \\
          decay && 0.75 \\
         \bottomrule
    \end{tabular}
    \caption{Hyperparameters used for the taggers.}
    \label{tab:tagger_hyperparameters}
\end{table}

Meanwhile, Table \ref{tab:tagger_acc} shows the performance of the taggers that we initially used to draw the error distributions and propose PoS tags with different levels of accuracy.

\begin{table}[htpb!]
    \centering
    \small
    \begin{tabular}{lc}
    \toprule
    & Tagger Accuracy \\
    \midrule
         Ancient Greek-Perseus & 90.14 \\
Armenian-ArmTDP & 92.22 \\
Basque-BDT & 94.74 \\
Bhojpuri-BHTB & 81.52 \\
Bulgarian-BTB & 98.26 \\
Estonian-EDT & 96.32 \\
Guajajara-TuDeT & 84.20 \\
Kiche-IU & 92.28 \\
Korean-Kaist & 94.34\\
Ligurian-GLT & 81.19 \\
Norwegian-Bokmål & 97.51\\
Persian-PerDT & 96.53\\
Skolt Sami-Giellagas & 80.03 \\
Turkish-BOUN & 91.31 \\
Vietnamese-VTB & 87.05 \\
Welsh-CCG & 91.76 \\
\bottomrule
    \end{tabular}
    \caption{Accuracy on test sets of biLSTM taggers trained for each treebank from which each error distribution was deduced and used to control accuracy for each treebank in experiments.}
    \label{tab:tagger_acc}
\end{table}

\section{Parsing hyperparameters}\label{app:parsers}
Table \ref{tab:parser_hyperparameters} details the hyperparameters used to train all the sequence labeling parsers evaluated in this work.

\begin{table}[htbp!]
\footnotesize
    \centering
        \tabcolsep=.25cm  
    \begin{tabular}{l p{1.5em} r}
    \toprule
    \textbf{Hyperparameter} & & \textbf{Value}\\
    \midrule
         Word embedding dimensions& & 100\\
         Character embedding dimensions && 30 \\
         Character hidden dimensions && 50 \\
         Hidden dimensions && 800 \\
         POS embedding dimension && 25 \\
         LSTM layers && 2 \\
         CNN laters && 4 \\
         Dropout && 0.5 \\
         Epochs&  & 50 \\
         training batch size && 8 \\
          learning rate&  & 0.02 \\
          momentum && 0.9 \\
          decay && 0.05 \\
         \bottomrule
    \end{tabular}
    \caption{Hyperparameters used for the sequence labeling parsers.}
    \label{tab:parser_hyperparameters}
\end{table}

\section{Additional results}\label{app:uas}

\begin{figure}[hbpt!]
    \centering
    \includegraphics[width=0.40\textwidth]{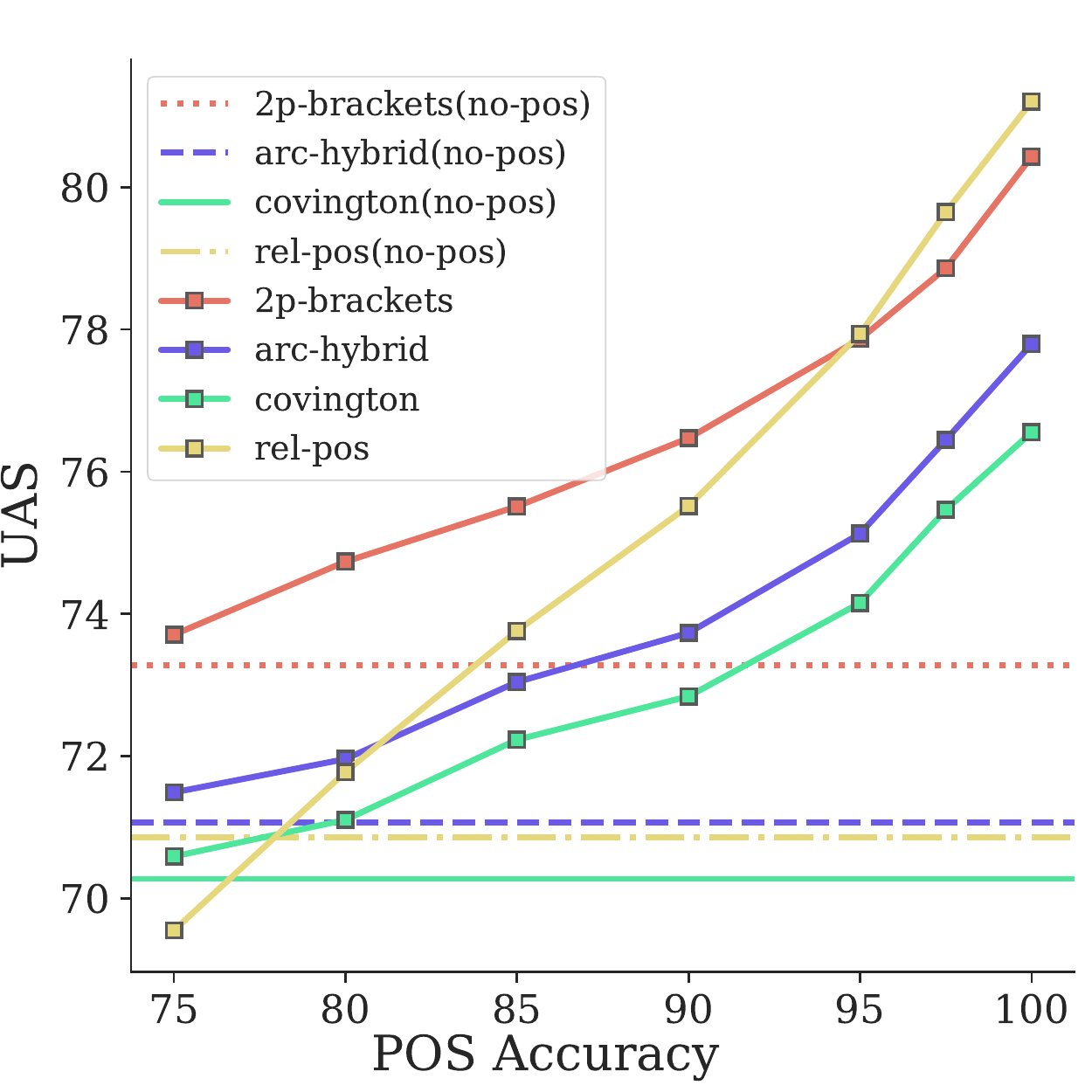}
    \caption{Average UAS across all treebanks against PoS tagging accuracies for different linearizations, compared to the no-tags baselines.}
    \label{fig:all_scores_uas}
\end{figure}

\begin{figure*}[t!]
    \centering
    \begin{subfigure}{0.4\textwidth}
    \includegraphics[width=\textwidth]{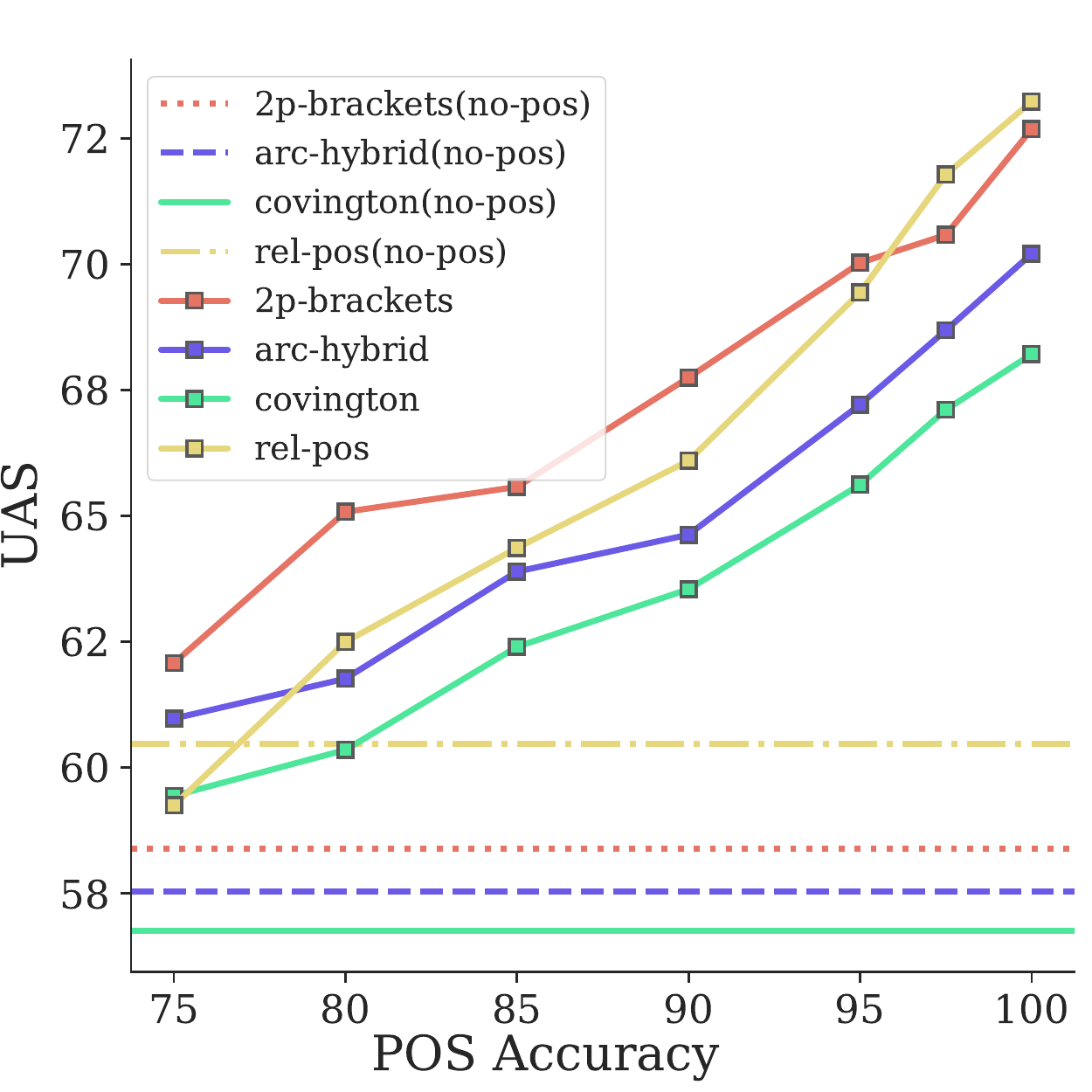}
    \caption{Low}
    \label{fig:low_split_uas}
    \end{subfigure}
   ~
    \begin{subfigure}{0.4\textwidth}
    \includegraphics[width=\textwidth]{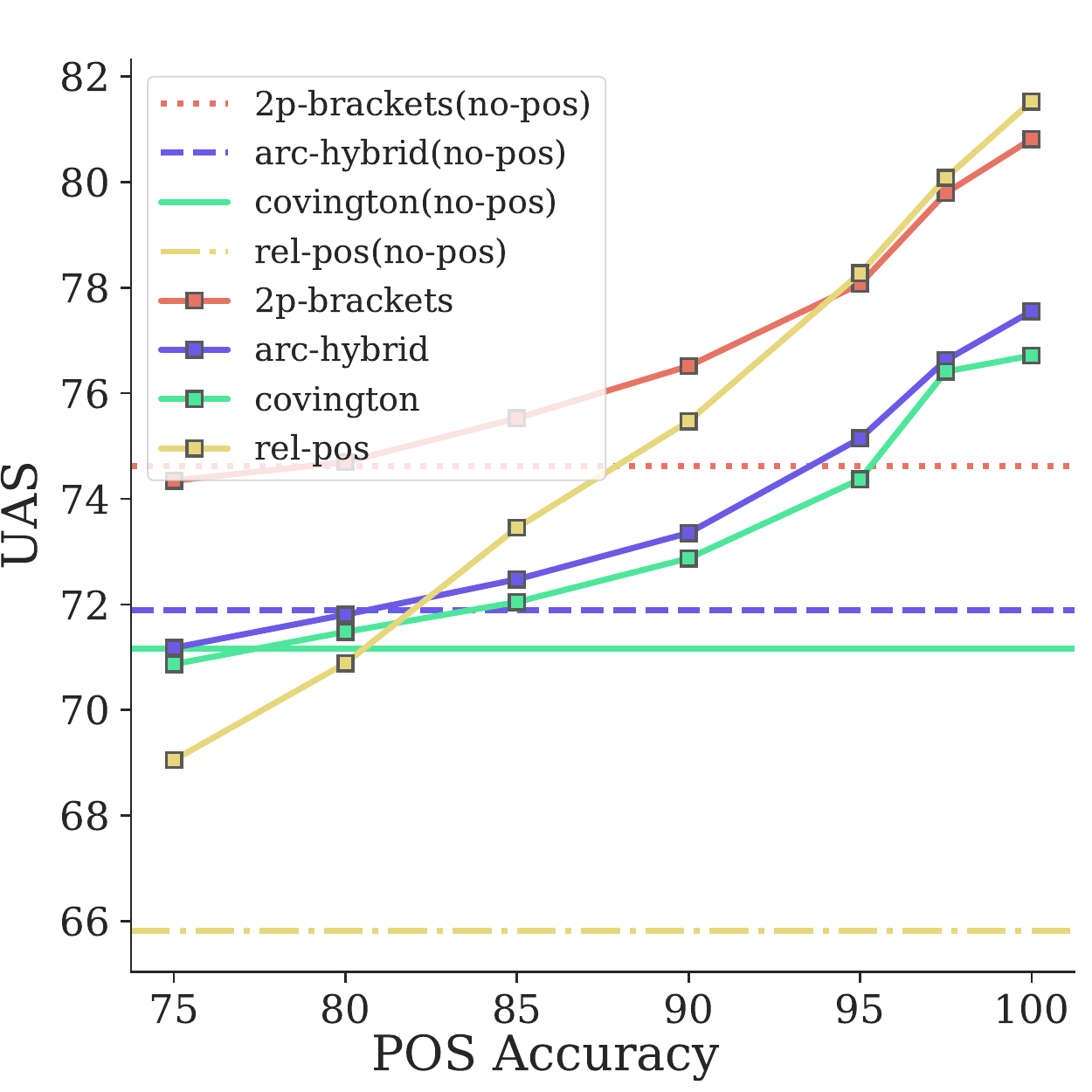}
    \caption{Mid}
    \label{fig:mid_split_uas}
    \end{subfigure}

    \begin{subfigure}{0.4\textwidth}
    \includegraphics[width=\textwidth]{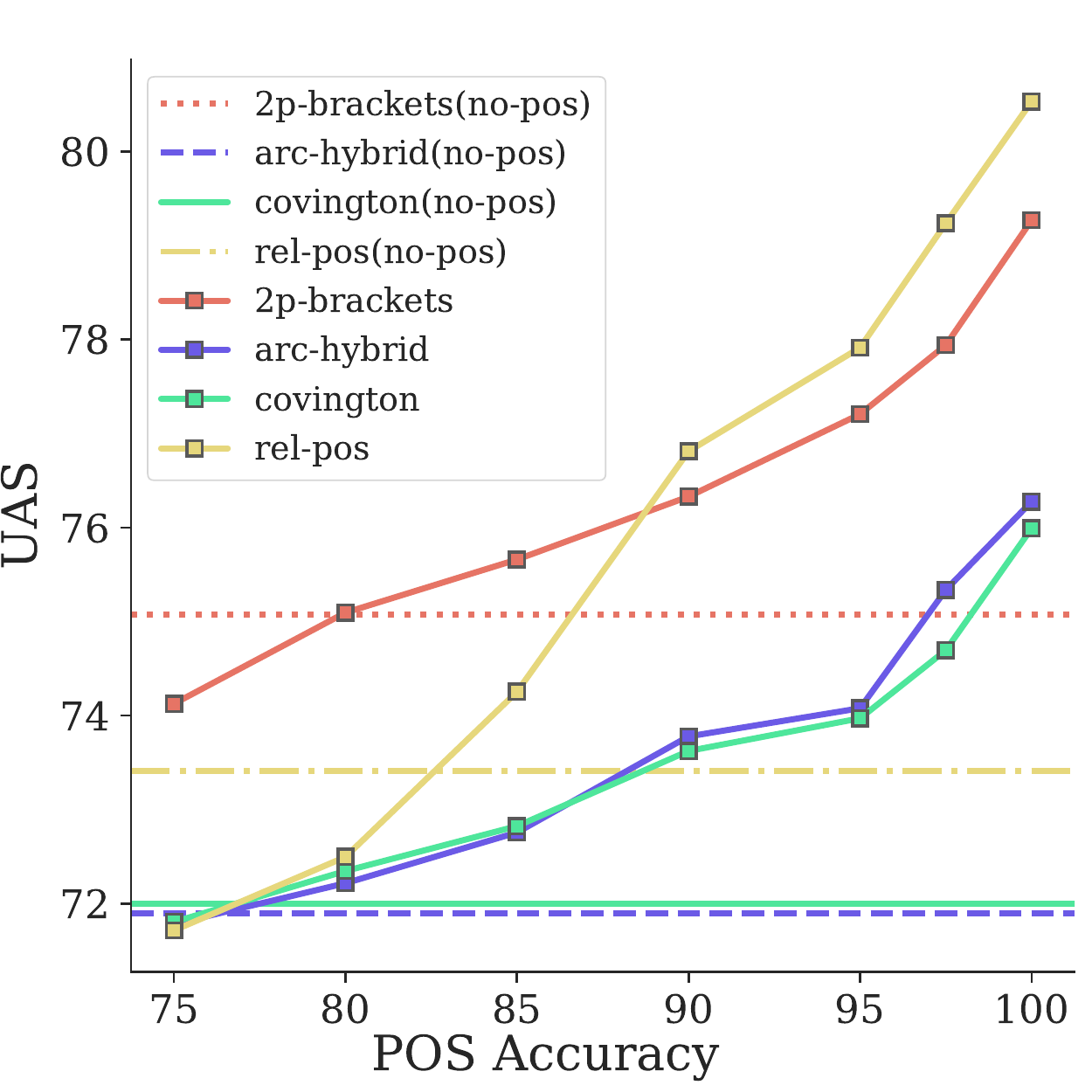}
    \caption{High}
    \label{fig:rich_split_uas}
    \end{subfigure}
    ~
    \begin{subfigure}{0.4\textwidth}
    \includegraphics[width=\textwidth]{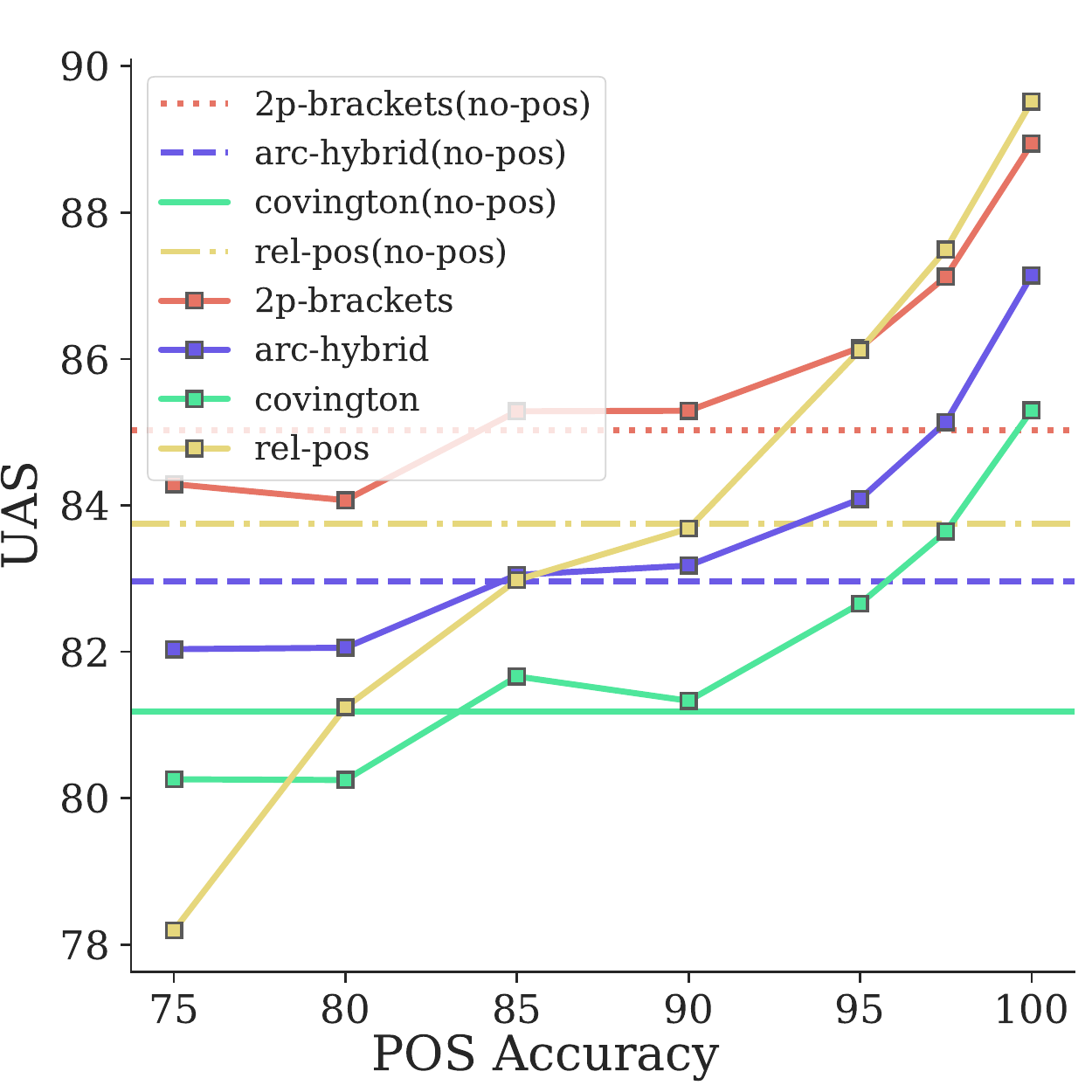}
    \caption{Very high}
    \label{fig:rich_split_uas}
    \end{subfigure}
    \caption{Average UAS for the (a) low-, (b) mid-, (c) high and (d) very high-resource subsets of treebanks for different PoS tagging accuracies and linearizations, compared to the no-tags baselines.}
    \label{fig:split_scores_uas}
\end{figure*}

\begin{figure}[hbpt!]
    \centering
    \includegraphics[width=0.40\textwidth]{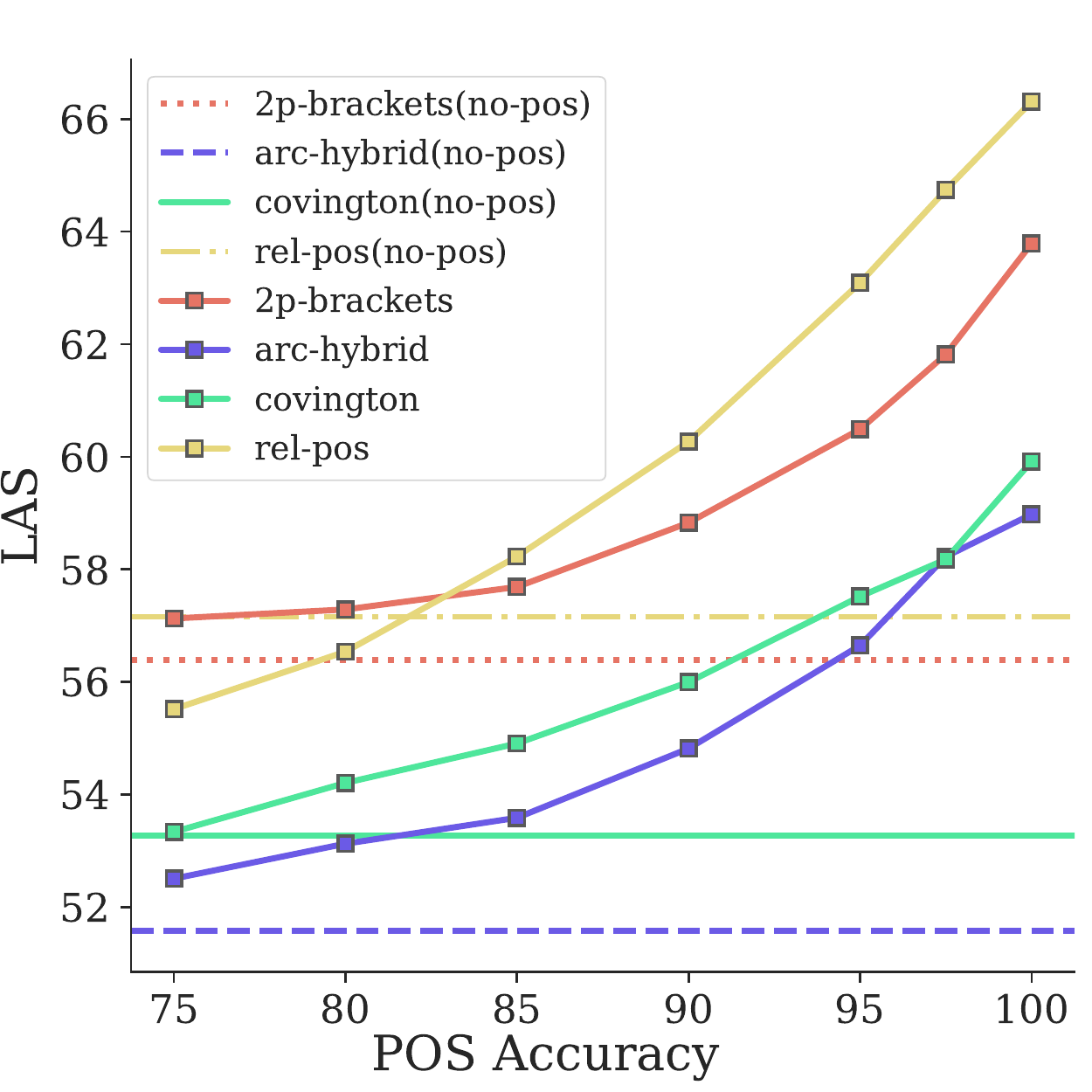}
    \caption{LAS against PoS tagging accuracies for different linearizations for the Ancient Greek\textsubscript{Perseus}, compared to the no-tags baselines.}
    \label{fig:ancient_greek_las}
\end{figure}

\begin{figure}[hbpt!]
    \centering
    \includegraphics[width=0.40\textwidth]{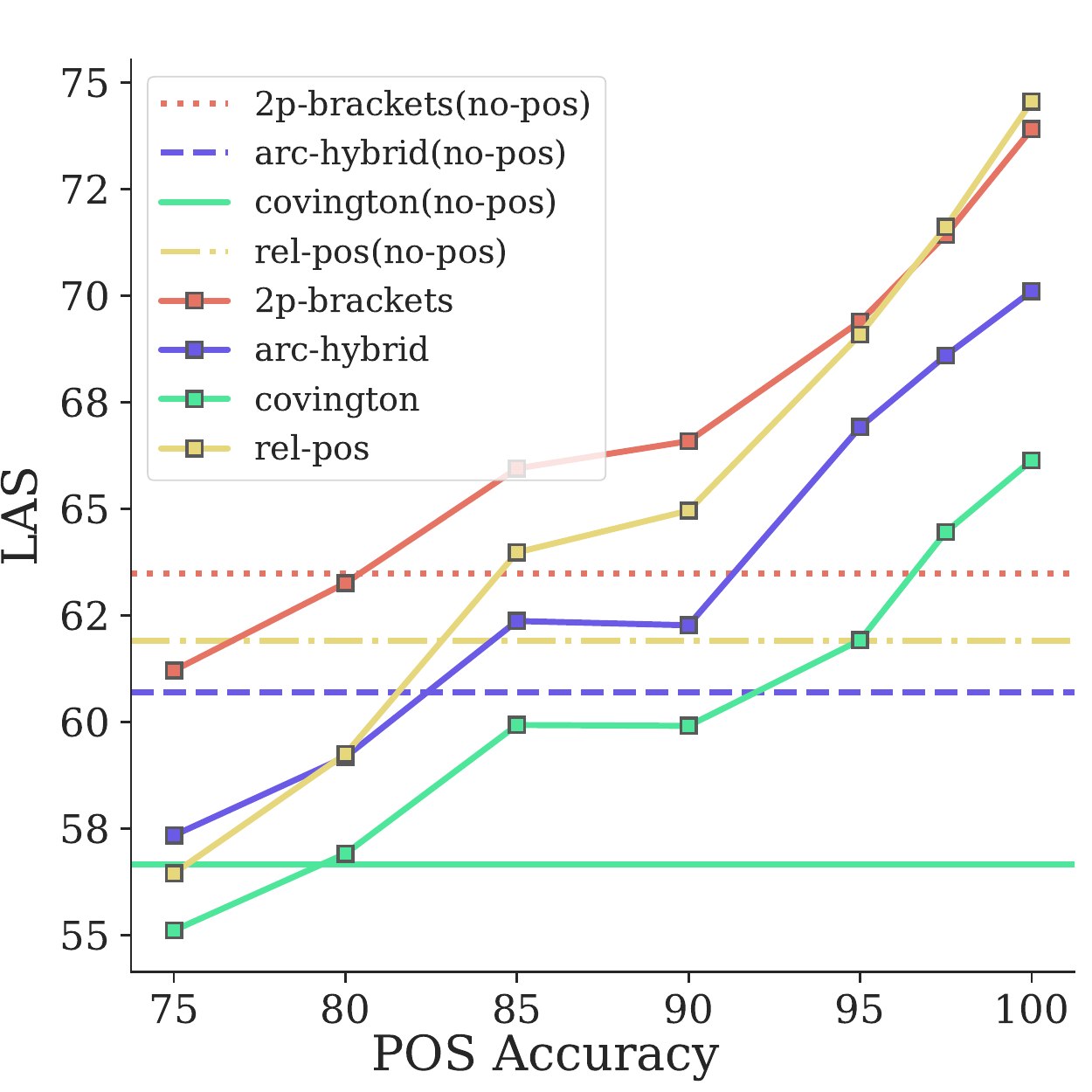}
    \caption{LAS against PoS tagging accuracies for different linearizations for the Armenian\textsubscript{ArmTDP}, compared to the no-tags baselines.}
    \label{fig:armenian_las}
\end{figure}

\begin{figure}[hbpt!]
    \centering
    \includegraphics[width=0.40\textwidth]{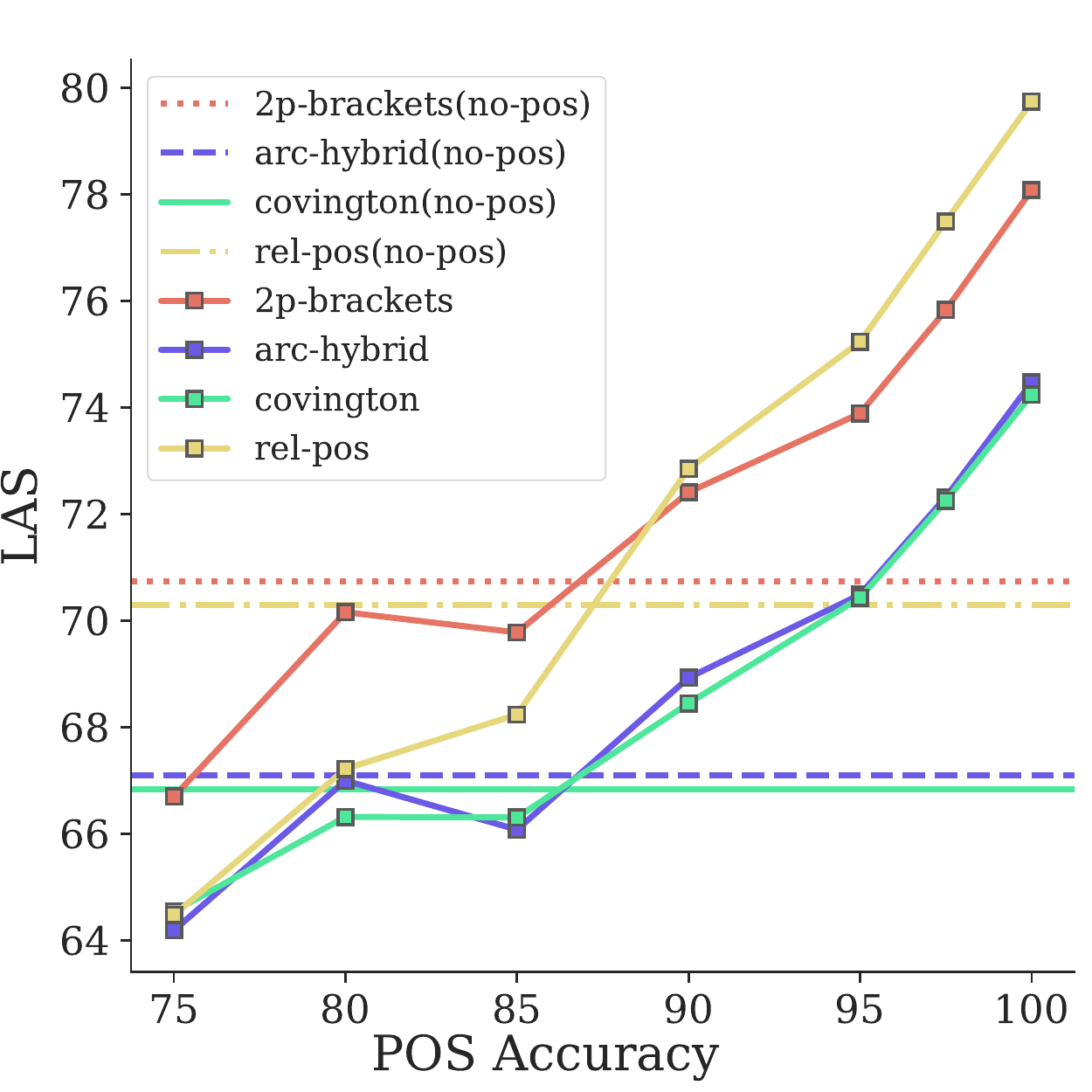}
    \caption{LAS against PoS tagging accuracies for different linearizations for the Basque\textsubscript{BDT}, compared to the no-tags baselines.}
    \label{fig:basque_las}
\end{figure}

\begin{figure}[hbpt!]
    \centering
    \includegraphics[width=0.40\textwidth]{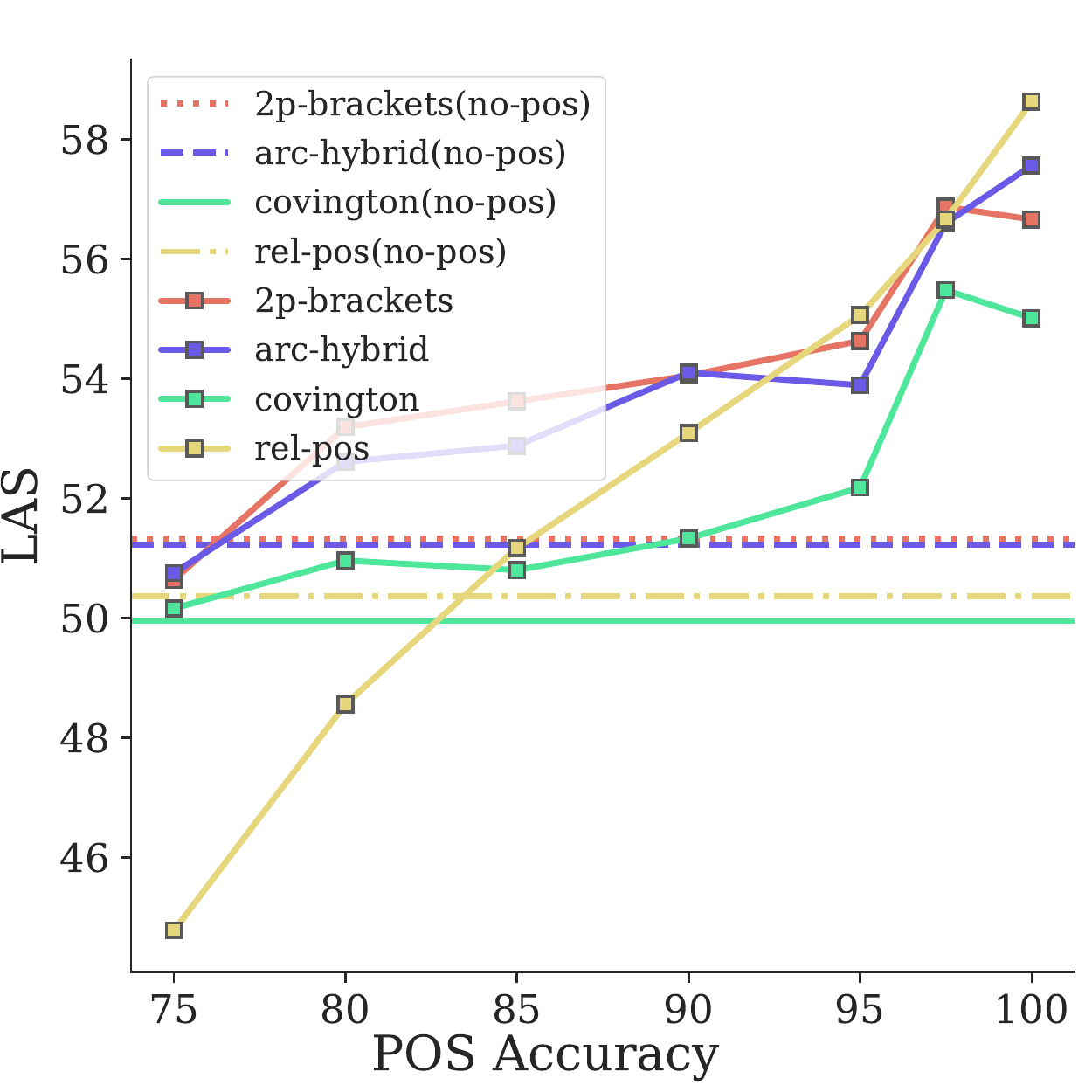}
    \caption{LAS against PoS tagging accuracies for different linearizations for the Bhojpuri\textsubscript{BHTB}, compared to the no-tags baselines.}
    \label{fig:bhojpuri_las}
\end{figure}

\begin{figure}[hbpt!]
    \centering
    \includegraphics[width=0.40\textwidth]{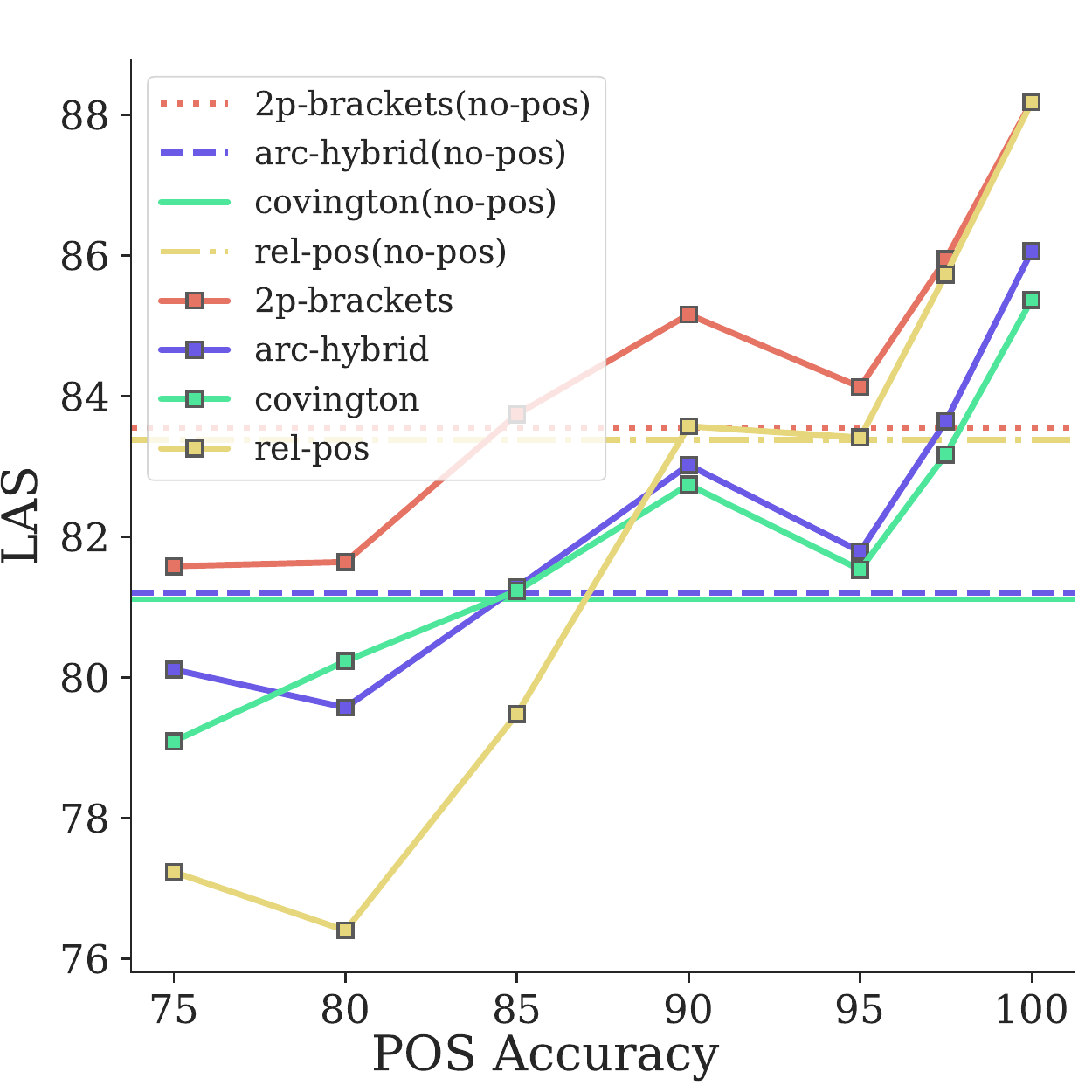}
    \caption{LAS against PoS tagging accuracies for different linearizations for the Bulgarian\textsubscript{BTB}, compared to the no-tags baselines.}
    \label{fig:bulgarian_las}
\end{figure}

\begin{figure}[hbpt!]
    \centering
    \includegraphics[width=0.40\textwidth]{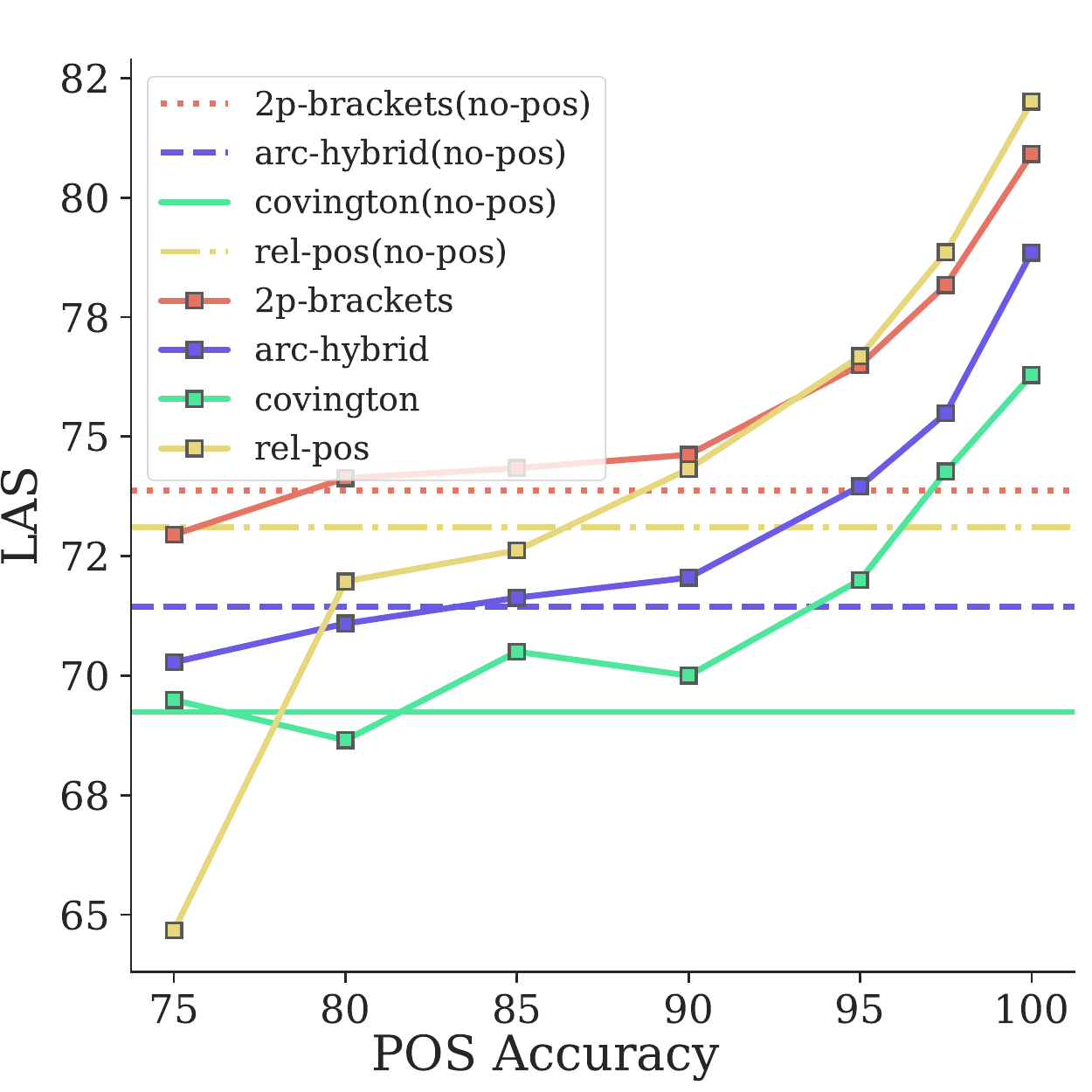}
    \caption{LAS against PoS tagging accuracies for different linearizations for the Estonian\textsubscript{EDT}, compared to the no-tags baselines.}
    \label{fig:estonian_las}
\end{figure}

\begin{figure}[hbpt!]
    \centering
    \includegraphics[width=0.40\textwidth]{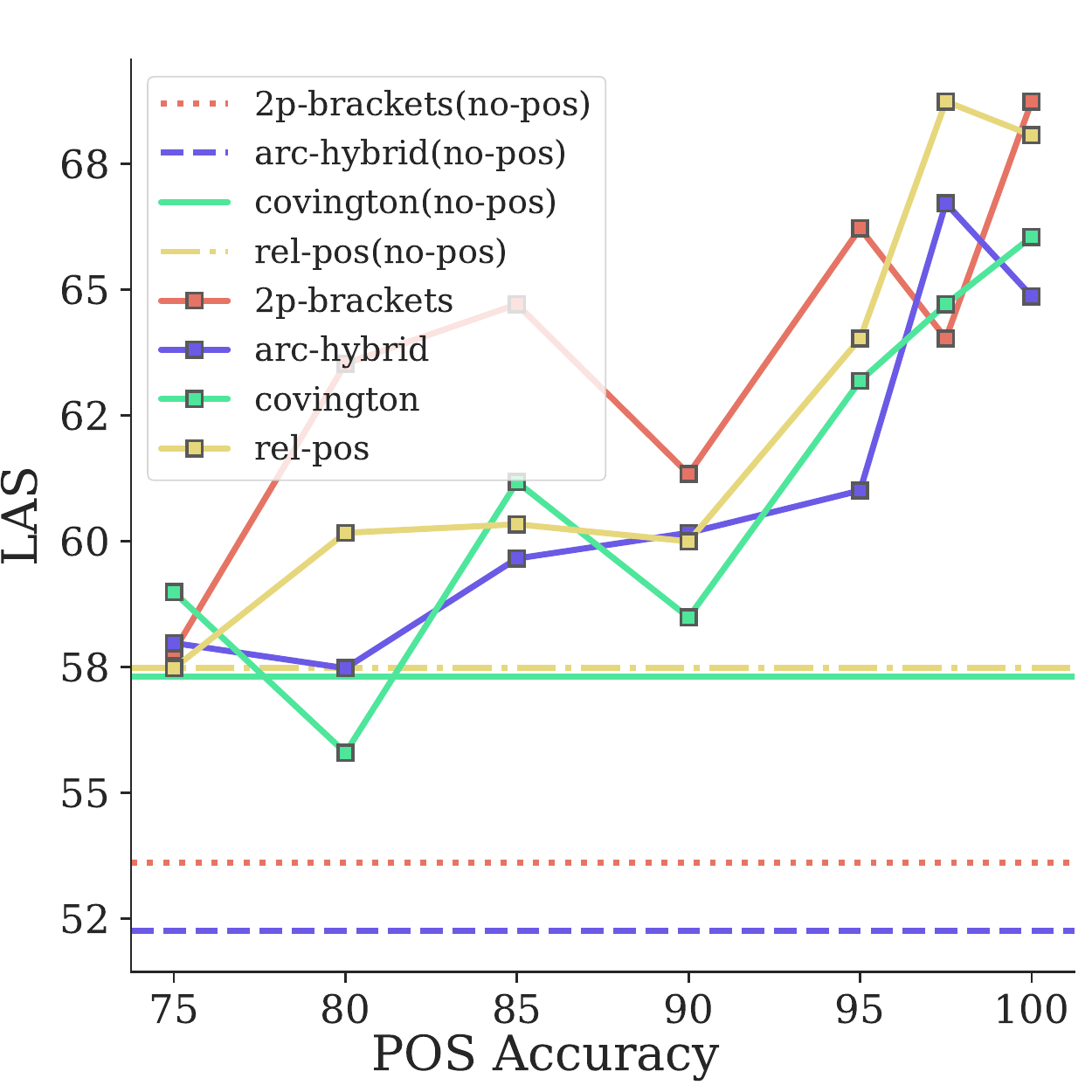}
    \caption{LAS against PoS tagging accuracies for different linearizations for the Guajajara\textsubscript{TuDeT}, compared to the no-tags baselines.}
    \label{fig:guajajara_las}
\end{figure}

\begin{figure}[hbpt!]
    \centering
    \includegraphics[width=0.40\textwidth]{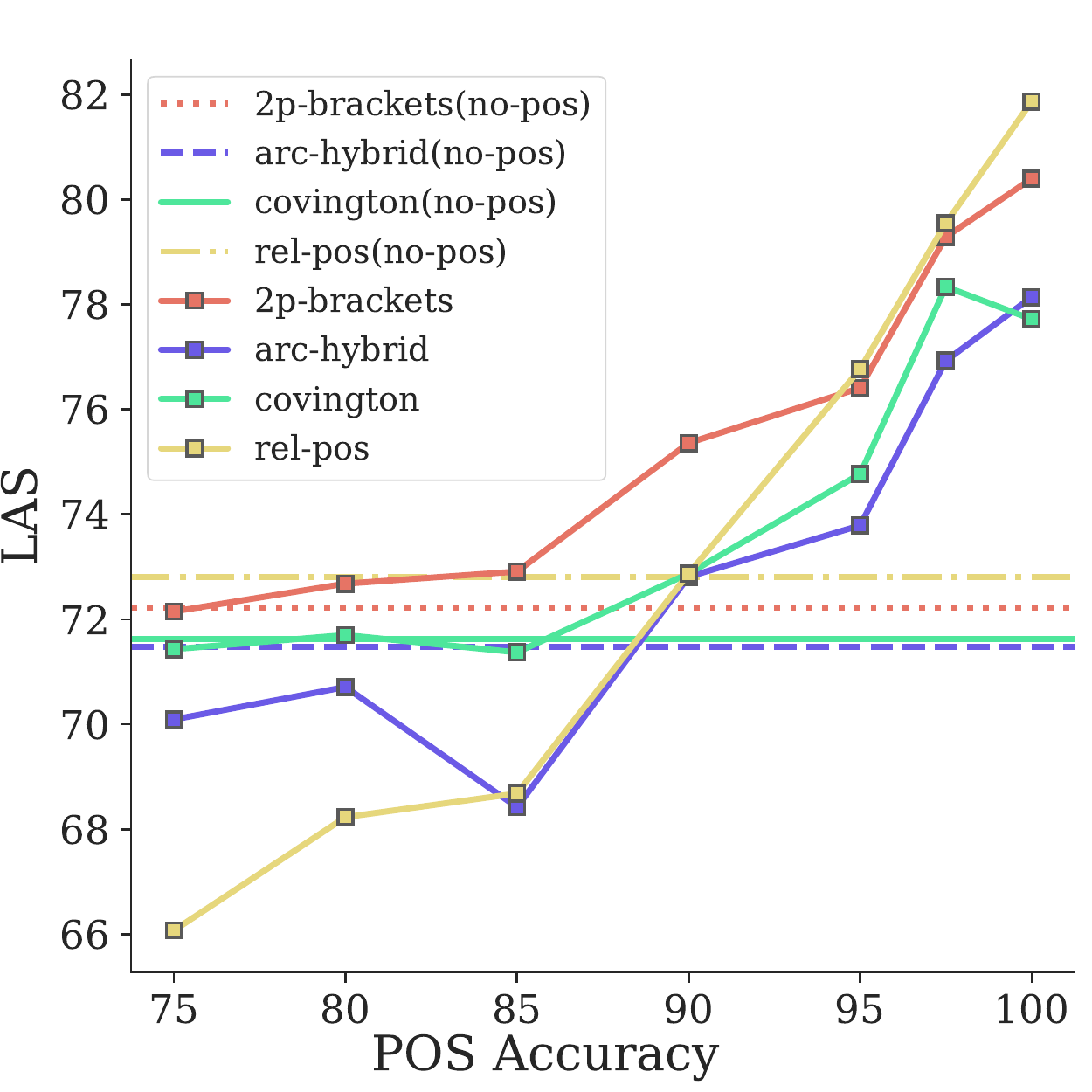}
    \caption{LAS against PoS tagging accuracies for different linearizations for the Kiche\textsubscript{IU}, compared to the no-tags baselines.}
    \label{fig:kiche_las}
\end{figure}

\begin{figure}[hbpt!]
    \centering
    \includegraphics[width=0.40\textwidth]{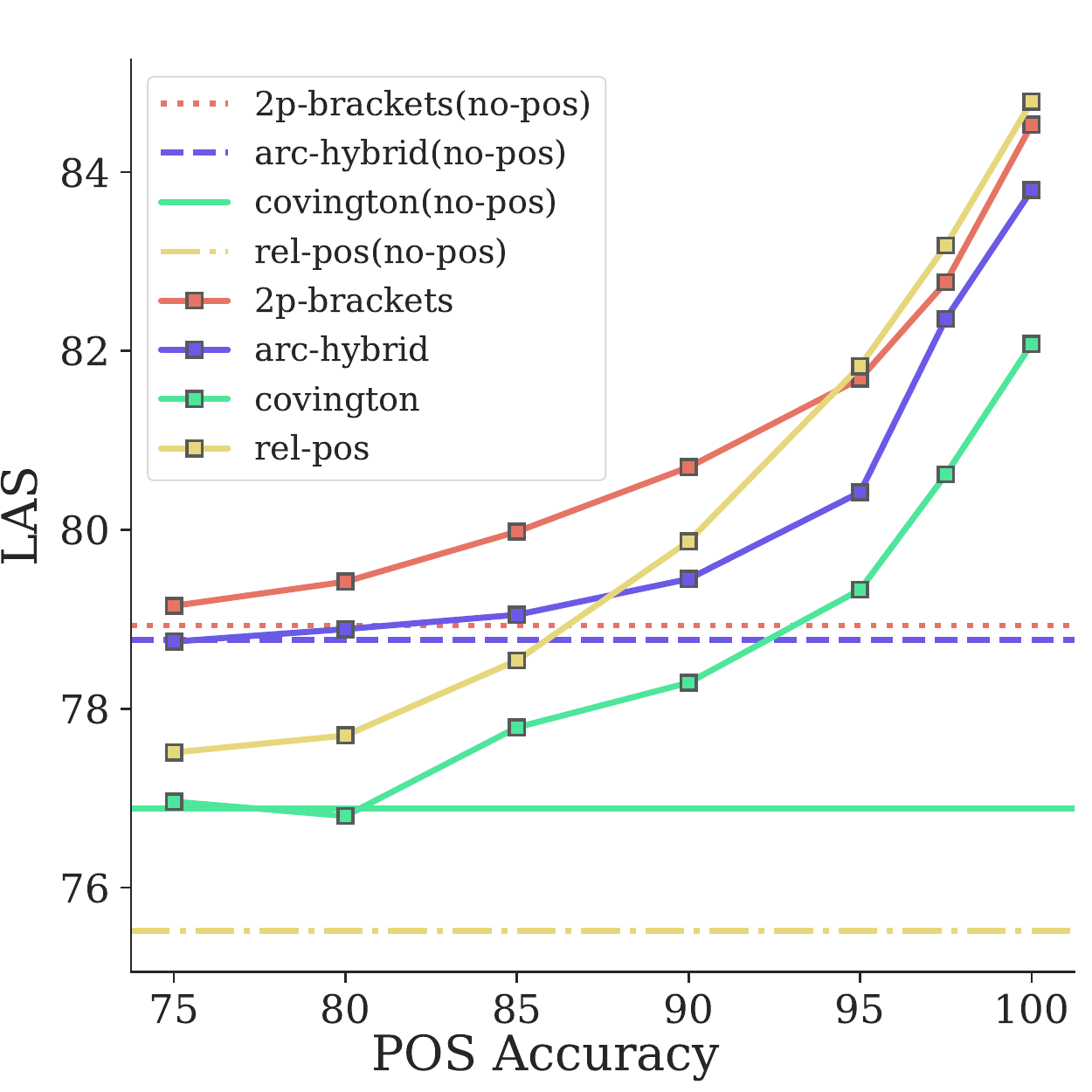}
    \caption{LAS against PoS tagging accuracies for different linearizations for the Korean\textsubscript{Kaist}, compared to the no-tags baselines.}
    \label{fig:korean_las}
\end{figure}

\begin{figure}[hbpt!]
    \centering
    \includegraphics[width=0.40\textwidth]{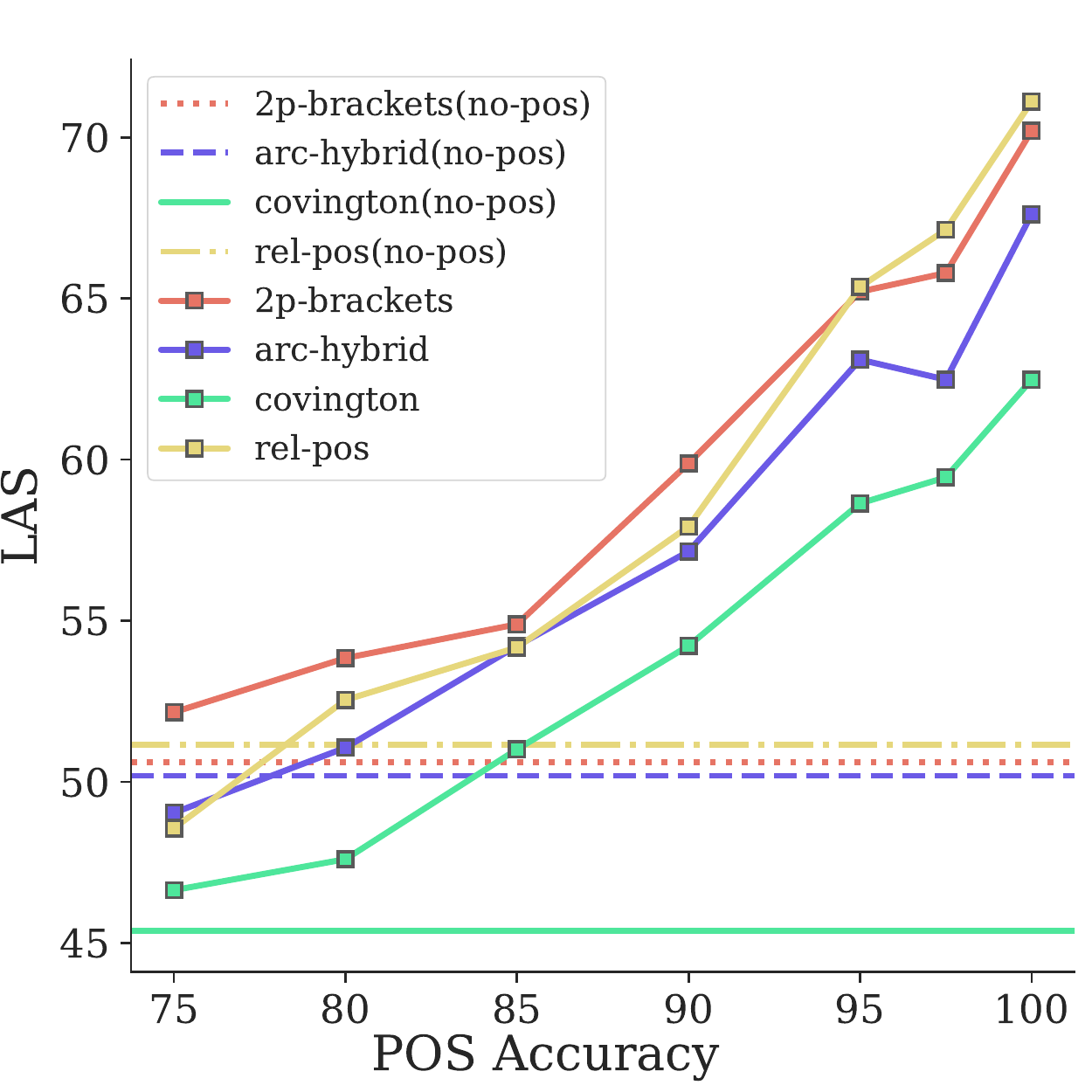}
    \caption{LAS against PoS tagging accuracies for different linearizations for the Ligurian\textsubscript{GLT}, compared to the no-tags baselines.}
    \label{fig:ligurian_las}
\end{figure}

\begin{figure}[hbpt!]
    \centering
    \includegraphics[width=0.40\textwidth]{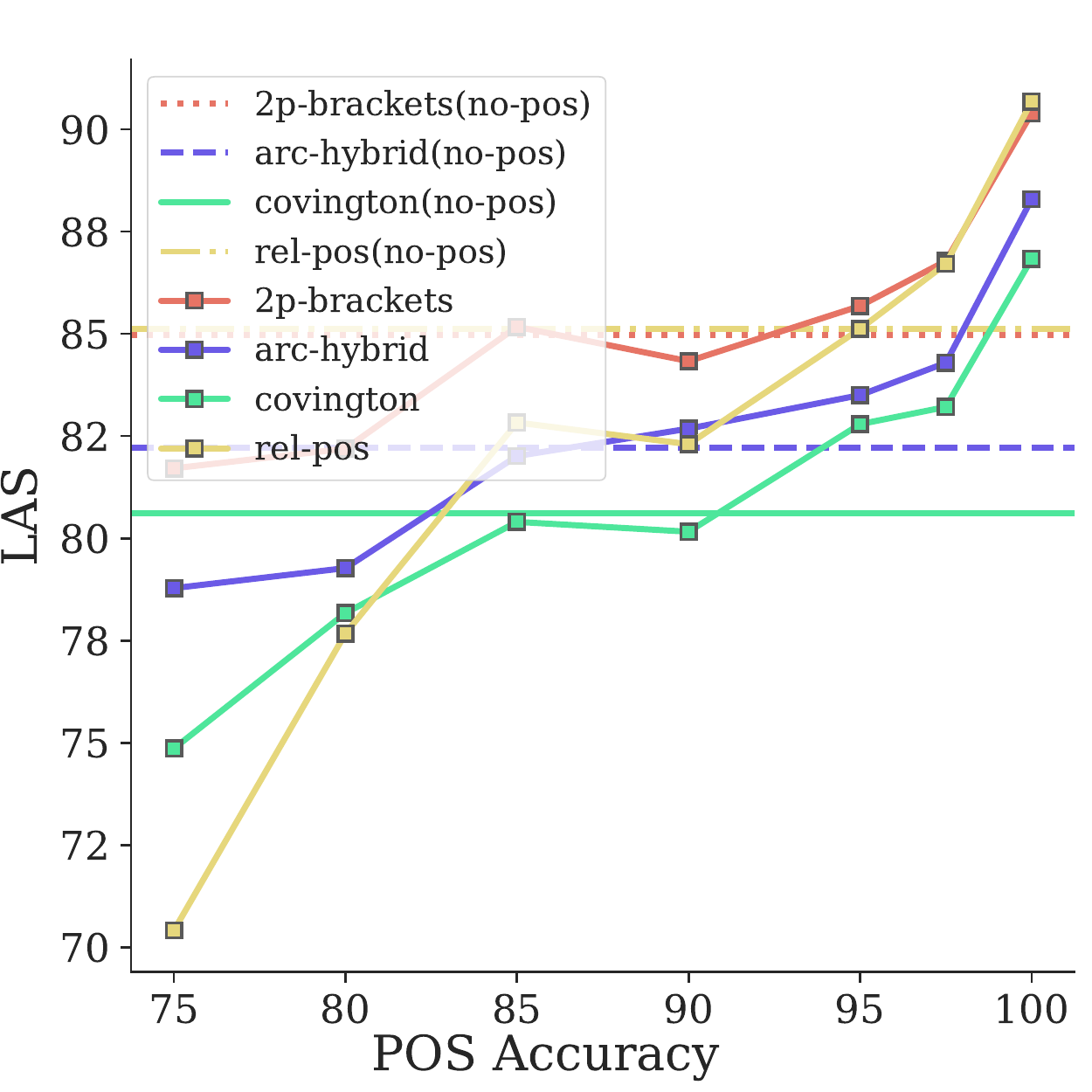}
    \caption{LAS against PoS tagging accuracies for different linearizations for the Norwegian\textsubscript{Bokmål}, compared to the no-tags baselines.}
    \label{fig:norwegian_las}
\end{figure}

\begin{figure}[hbpt!]
    \centering
    \includegraphics[width=0.40\textwidth]{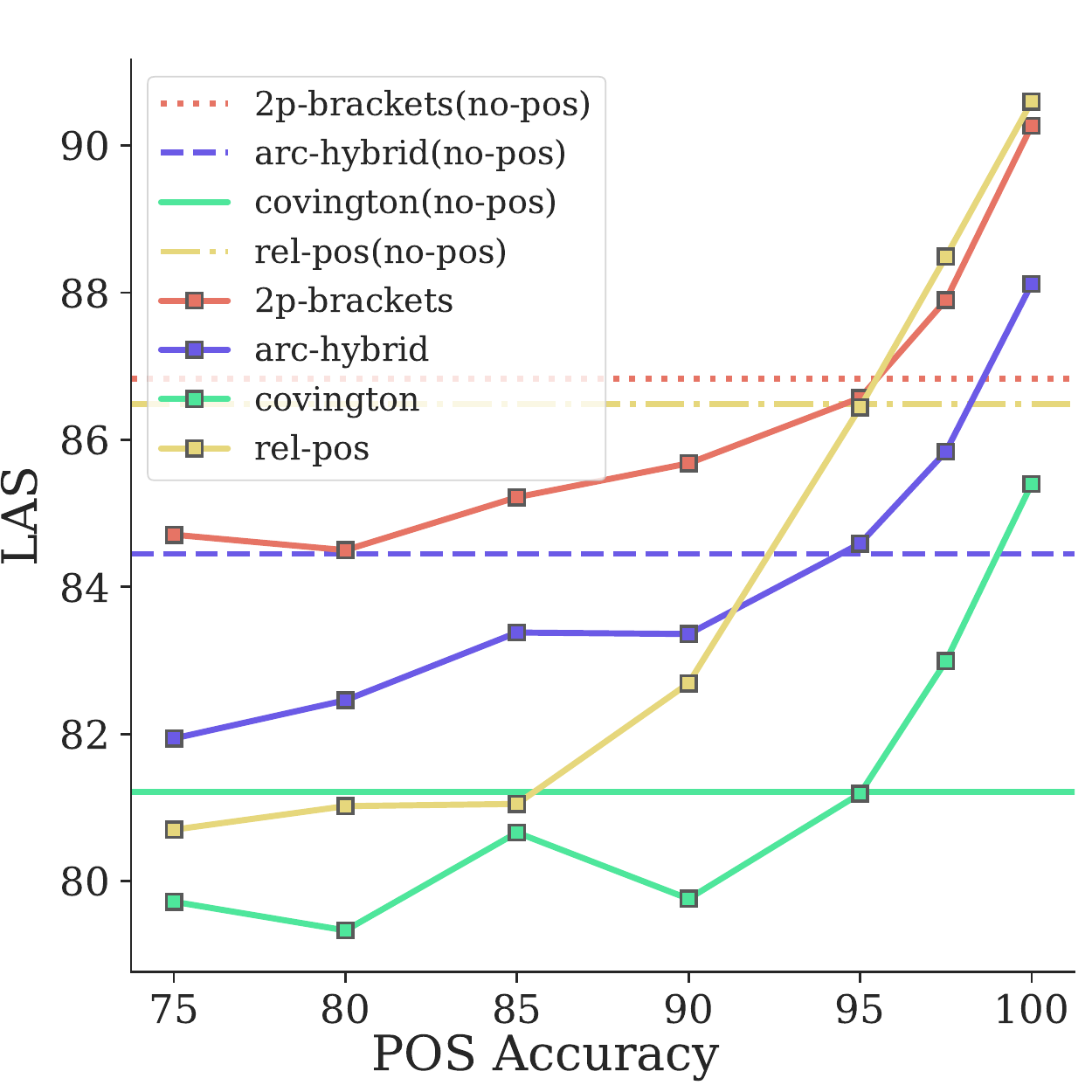}
    \caption{LAS against PoS tagging accuracies for different linearizations for the Persian\textsubscript{PerDT}, compared to the no-tags baselines.}
    \label{fig:persian_las}
\end{figure}

\begin{figure}[hbpt!]
    \centering
    \includegraphics[width=0.40\textwidth]{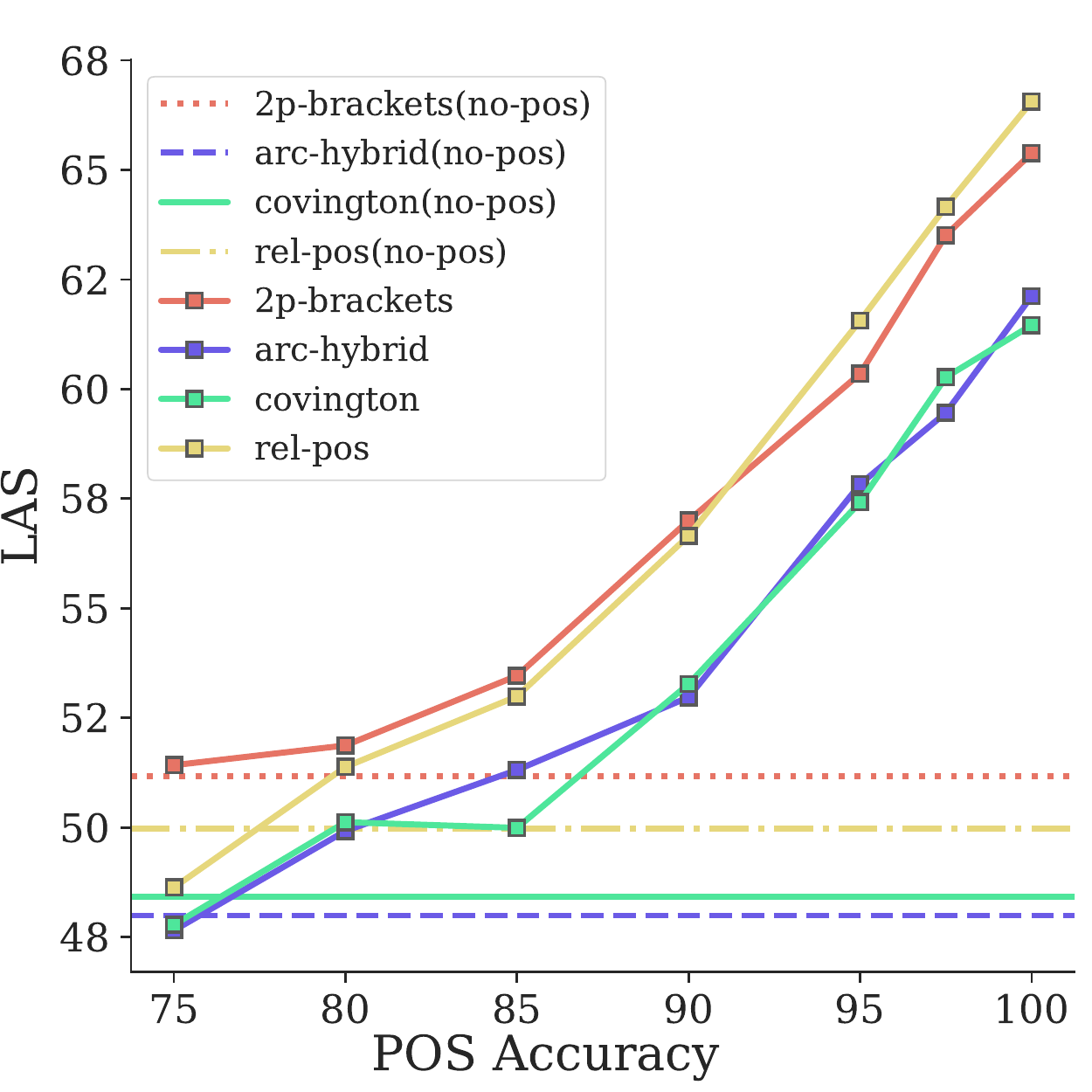}
    \caption{LAS against PoS tagging accuracies for different linearizations for the Vietnamese\textsubscript{VTB}, compared to the no-tags baselines.}
    \label{fig:vietnamese_las}
\end{figure}

\begin{figure}[hbpt!]
    \centering
    \includegraphics[width=0.40\textwidth]{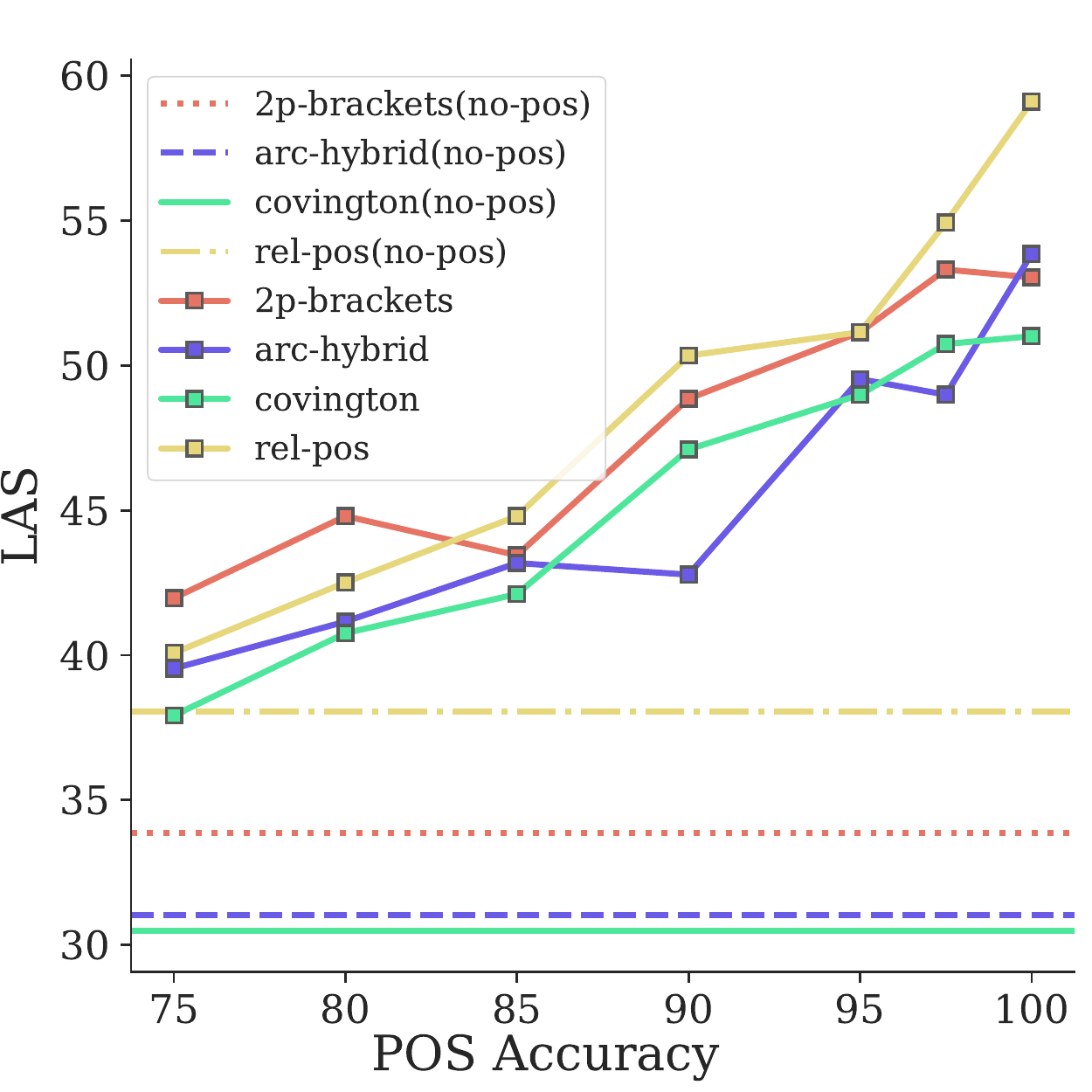}
    \caption{LAS against PoS tagging accuracies for different linearizations for the Skolt Sami\textsubscript{Giellagas}, compared to the no-tags baselines.}
    \label{fig:sami_las}
\end{figure}

\begin{figure}[hbpt!]
    \centering
    \includegraphics[width=0.40\textwidth]{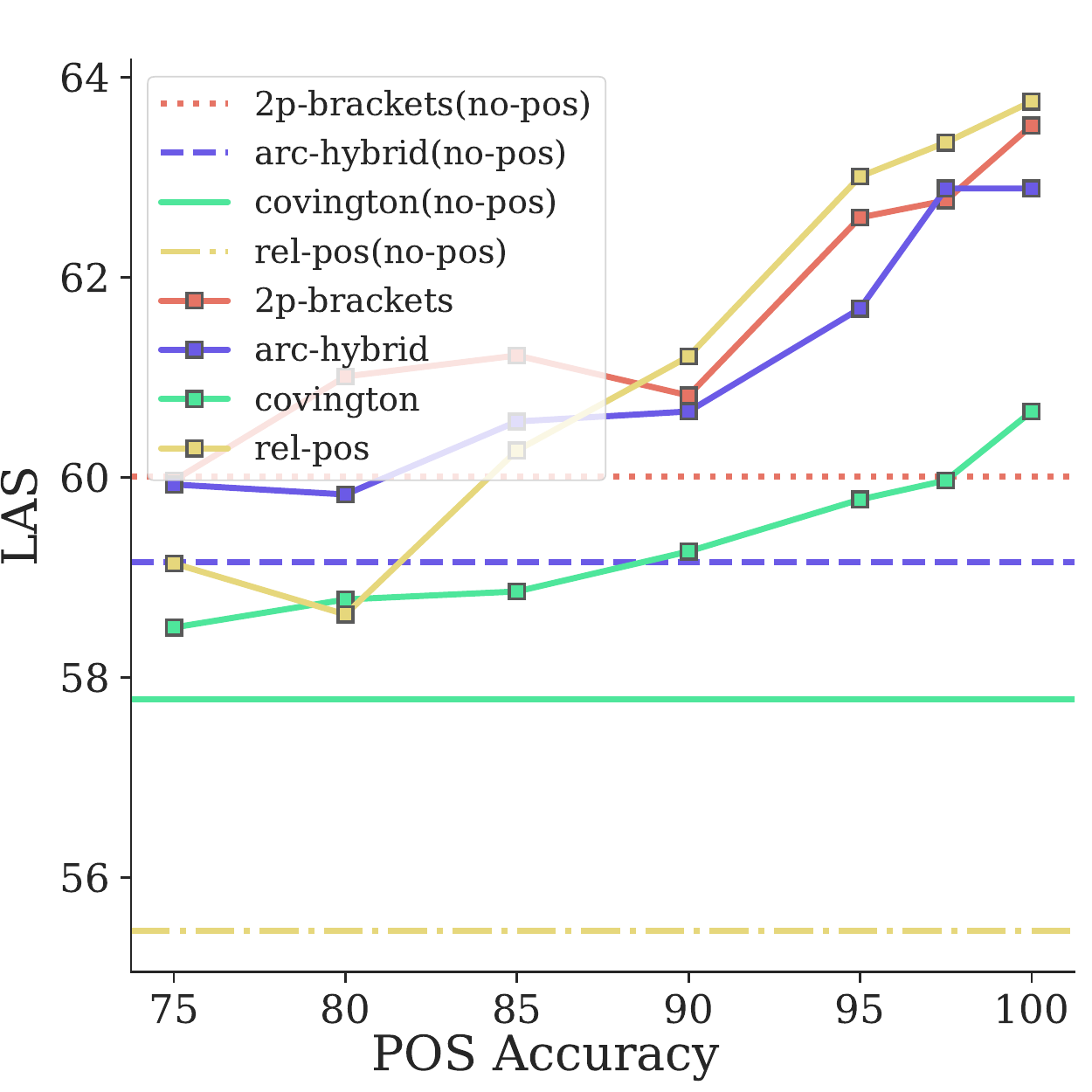}
    \caption{LAS against PoS tagging accuracies for different linearizations for the Turkish\textsubscript{BOUN}, compared to the no-tags baselines.}
    \label{fig:turkish_las}
\end{figure}

\begin{figure}[hbpt!]
    \centering
    \includegraphics[width=0.40\textwidth]{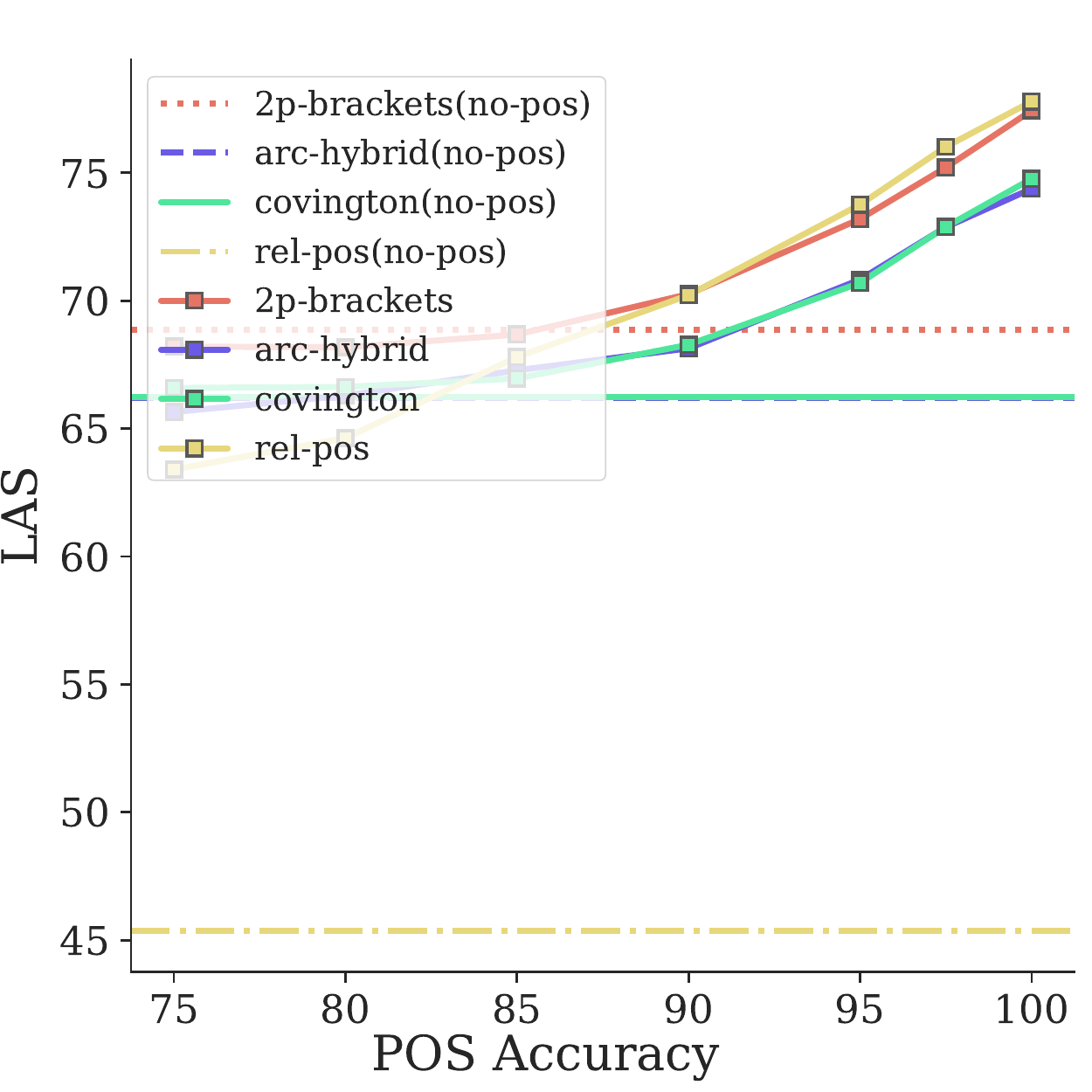}
    \caption{LAS against PoS tagging accuracies for different linearizations for the Welsh\textsubscript{CCG}, compared to the no-tags baselines.}
    \label{fig:welsh_las}
\end{figure}

Figures \ref{fig:all_scores_uas} and \ref{fig:split_scores_uas} shows the UAS results complementing the LAS results reported in \S \ref{sec:experiments} (in Figures \ref{fig:all_scores} and \ref{fig:split_scores}, respectively). Figures from \ref{fig:ancient_greek_las} to \ref{fig:welsh_las} show the LAS results for each treebank.

\end{document}